\documentclass[11pt]{article}

\usepackage[preprint]{acl}

\usepackage{times}
\usepackage{latexsym}

\usepackage[T1]{fontenc}
\usepackage[utf8]{inputenc}

\usepackage{microtype}

\usepackage{inconsolata}

\usepackage{graphicx}

\usepackage{natbib}

\makeatletter
\let\oldlt\longtable
\let\endoldlt\endlongtable
\def\longtable{\@ifnextchar[\longtable@i \longtable@ii}
\def\longtable@i[#1]{\begin{figure}[t]
\onecolumn
\begin{minipage}{0.5\textwidth}
\oldlt[#1]
}
\def\longtable@ii{\begin{figure}[t]
\onecolumn
\begin{minipage}{0.5\textwidth}
\oldlt
}
\def\endlongtable{\endoldlt
\end{minipage}
\twocolumn
\end{figure}}
\makeatother

\newcommand\blfootnote[1]{%
  \begingroup
  \renewcommand\thefootnote{}\footnote{#1}%
  \addtocounter{footnote}{-1}%
  \endgroup
}

\title{TelcoLM: collecting data, adapting, and benchmarking language models for the telecommunication domain}

\author{Camille Barboule$^*$ ~~ Viet-Phi Huynh$^*$ ~~ Adrien Bufort\\
        \textbf{Yoan Chabot} ~~ \textbf{Géraldine Damnati} ~~ \textbf{Gwénolé Lecorvé}\\
        Orange, France\\
        \texttt{\{camille.barboule,vietphi.huynh,adrien.bufort,}\\
        \texttt{yoan.chabot,geraldine.damnati,gwenole.lecorve\}@orange.com}
        }

\begin{document}

\maketitle

\begin{abstract}
Despite outstanding processes in many tasks, Large Language Models (LLMs) still lack accuracy when dealing with highly technical domains.
Especially, telecommunications (telco) is a particularly challenging domain due the large amount of lexical, semantic and conceptual peculiarities. Yet, this domain holds many valuable use cases, directly linked to industrial needs.
Hence, this paper studies how LLMs can be adapted to the telco domain. It reports our effort to (i) collect a massive corpus of domain-specific data (800M tokens, 80K instructions), (ii) perform adaptation using various methodologies, and (iii) benchmark them against larger generalist models in downstream tasks that require extensive knowledge of telecommunications.
Our experiments on Llama-2-7b show that domain-adapted models can challenge the large generalist models. They also suggest that adaptation can be restricted to a unique instruction-tuning step, dicarding the need for any fine-tuning on raw texts beforehand.
\end{abstract}

\blfootnote{$^*$ Main authors with equal contribution. Work achieved in March, 2024. Released in December, 2024.}

\section{Introduction}
\label{sec:introduction}

Large Language Models like GPT-4~\cite{openai2023gpt4}, PaLM~\cite{Chowdhery2022palm}, LLaMA~\cite{Touvron2023llama}, Falcon~\cite{ZXhang2023falcon} or Mistral~\cite{jiang2023mistral} perform very well on a wide range of tasks, on a wide range of tasks, both when the knowledge required is general and when it relates to various domains.
Still, their application to highly specialized tasks, requiring advanced and up-to-date knowledge, raises a number of challenges~\cite{zhao2023domain}:
general and frequently discussed topics tend to be disproportionately represented in their corpus~\cite{Penedo2023refinedweb}, whereas highly domain-specific topics tend to be underrepresented, which inevitably leads to challenges in effectively learning them for domain-specific tasks.
Additionally, many domain-specific knowledge resources are proprietary assets, critical to an organization's competitive edge, and cannot be readily shared with general-purpose LLMs.

\begin{figure}
    \centering
    \includegraphics[scale=0.75]{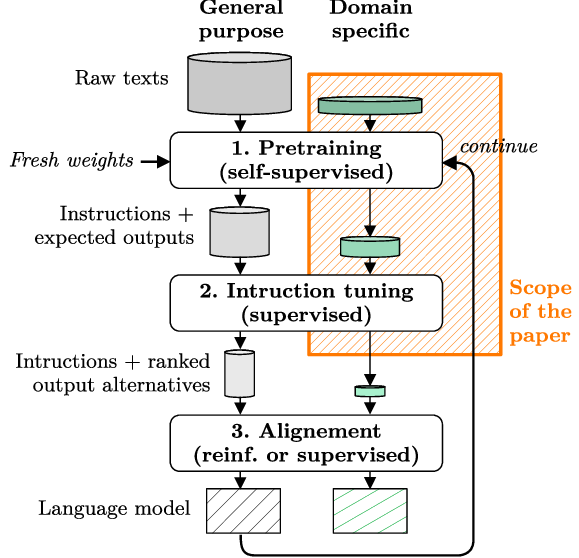}
    \caption{Steps for LM training and adaptation.}
    \label{fig:adaptation_process}
\end{figure}

Adapting a language model to a target domain, called \textit{domain adaptation}, is a deeply explored in the literature to solve these problems~\cite{zhao2023domain}.
Basically, this can be performed by rerunning the training steps of a baseline general-purpose LM on domain-specific data. As illustrated in Figure~\ref{fig:adaptation_process}, these steps consists in (1) a pretraining step in a self-supervised manner on raw texts, (2) an instruction-tuning step on supervised datasets of instruction-output pairs, and (3) an alignement step to help the model generate texts which fits the behavioral expectations of the users.

Among domains of interests, several studies~\cite{maatouk2023large,bariah2023large,bariah2023understanding} emphasize the importance of undertaking the adaptation of language models for the telecom industry because this sector not only exhibits a large amount of lexical, semantic and concept specificities~\cite{bariah2023understanding,holm2021bidirectional}, but it is also a significant provider of natural language processing tasks, from understanding technical documents to incident resolution or network modeling~\cite{maatouk2023large}.
Furthermore, in industrial applications, adapting reasonable-size models can significantly reduce computational costs, and ease deployments in environments with limited resources.

In this paper, we present experimental work to design the most efficient approach for adaptation to the telco domain. 
In details, our contributions concern: 
\begin{itemize}
    \item Data: We expose our collecting process to gather raw texts (800M tokens) and instructions (80k instructions) on the telco domain.
    \item Evaluation: We present a benchmark of telco-specific downstream tasks, combined with general-purpose evaluation to measure potential performance drops.
    \item Adaptation: We report a detailed comparison of various adaptation approaches by disabling or enabling some steps or some data sources.
\end{itemize}
The baseline model is Llama-2-7B~\cite{Touvron2023llama}.
The key conclusions are that the pretraining step can be skipped, and the best adapted models compete with GPT3.5.

% Such an adaptation can be processed with Domain Adaptive Pretraining (DAPT)~\cite{gururangan2020domainadaptation}, which consists in continuing the pre-training process on domain-specific corpora. This method has demonstrated its effectiveness on various Language Models~\cite{elboukkouri2022dapt,electra,gururangan2020domainadaptation}. 
% However, if this process performed well on small-sized language models,~\cite{readingcomprehension2023} revealed that continued training on the raw corpora results in a drastic drop in prompting performance of LLMs, making them unusable on downstream tasks. 
% This is why many domain adaptation works combine this DAPT approach with instruction-tuning~\cite{chipnemo,K2,foodgpt,Zhongjing}, allowing the model to leverage the knowledge acquired in DAPT on domain-specific downstream tasks. 
% However, this 2-steps approach for domain-adaptation is costly and complex. This is why recent work has focused on a single-step process for domain adaptation of LLMs, consisting in transforming the information from the raw text into instruction-like text~\cite{readingcomprehension2023,huatogpt2}. These approaches are similar to the Task-Adaptive Pretraining (TAPT) approach introduced by~\cite{gururangan2020domainadaptation}, consisting in continuing the pretraining process on domain-specific-task data. Since the recent method are using domain-instruction data rather than domain-task-specific data, we will call this process Instruction-Adaptive Pretraining (IAPT) in the rest of the paper.

% Paper outline
To achieve these goals, we organize the paper as follows:
Section~\ref{sec:related_work} reviews the current methods' for specialized domain adaptation.
Section~\ref{sec:methodology} then describes the methodology used, including the data collection, how we process this domain adaptation and how we evaluate it.
Experimental results including evaluation on several tasks related to telecom are then presented in Section~\ref{sec:evaluation}.
After discussing the performance of the various fine-tuning methods, we conclude the paper in Section~\ref{sec:conclusion}.

% ----------------------------------------------
% ----------------------------------------------
% ----------------------------------------------

\section{Related Work}
\label{sec:related_work}
In this section, we examine the various approaches employed for addressing domain adaptation in the literature. 
Prominent among these are the LoRA~\cite{hu2021lora}, QLoRA~\cite{dettmers2023qlora}, and ReLoRA~\cite{lialin2023stack} fine-tuning methods. 

\textbf{Domain adaptation without modifying the model's weights}\newline
Domain adaptation can be simply adressed using a retriever~\cite{Guu2020retriever} which gives to the model access to a wide range of external documents. A retrieval-augmented LM allows the model to have access, in the prompt context, to an external Knowledge Base (a new corpus)~\cite{Piktus2021RAG}. This retriever enables the model to access and focus on documents within an extensive corpus, such as domain internal knowledge, giving to the model access to many sources it hasn't seen during the pretraining and fine-tuning phases. However, adding retrieval augmentation to a language model alone is insufficient for incorporating knowledge into the model. \cite{Zhang2021retrieverLimits} demonstrates that even though retrieval-augmented models were capable of updating certain knowledge when the retrieval corpus was swapped, the performance of the retrieval-augmented language model on new knowledge-related questions is notably low. In contrast, the model performs significantly better on questions related to the knowledge present in the original training corpus. The researchers also noted that substantial improvements in handling new knowledge questions only occurred after fine-tuning the model with fresh data. This indicates that merely updating the corpora from which models retrieve passages is not enough to effectively integrate knowledge into a language model. The reason behind this behavior is elucidated by the memorization process that occurs in LLMs during their pretraining phase. This memorization hinders the model's ability to effectively handle knowledge sourced from external documents via a retriever~\cite{Longpre2022retrieverLimits}. During the pretraining phase, LLMs acquire knowledge through memorization, enabling them to generate competitive results solely based on their own parametric knowledge, without the need for access to relevant documents, but~\cite{Longpre2022retrieverLimits} demonstrates that this memorization behavior contradicts the expectation that the model should provide responses consistent with the information it retrieves, thereby reducing the system's interpretability. Most problematically, this memorization behavior severely restricts the model's ability to generalize to new knowledge not present in its training data. Furthermore,~\cite{Longpre2022retrieverLimits} highlights that the extent of memorization during the pretraining phase increases proportionally with the model's size, meaning that larger models exhibit a greater degree of memorization, and so less ability to generalize to new knowledge not present in its training data. Based on these demonstrations, the Retrieval-augmented methods are not suited to domain-adaptation, requiring lot of new and technical knowledge injection.

\textbf{Domain adaptation by adding knowledge in model's weights can be addressed through two methods: either training a model from scratch on domain-specific data or retraining a pretrained model on domain data (DAPT)}\newline
Training a model from scratch on domain-specific data enables the training of a tokenizer tailored to that specific domain, rather than retraining the model using a domain-specific corpus, which is limited by the use of a pretrained tokenizer with a general vocabulary. \cite{elboukkouri2022dapt} highlighted that retraining a general model on a specialized corpus, such as biomedical text, ultimately yields better results and is also significantly more cost-effective than training a model from scratch. This advantage becomes even more pronounced when working with LLMs. Also,~\cite{holm2021bidirectional} points out that DAPT is less sensitive to data quality, making it less likely to fail and less costly than pre-training from scratch. However, the study has also shown that DAPT cannot perform well when the discrepancy between the specialized domain and the general domain is too
large. But for the telco domain, the study shows that DAPT performs well.

\textbf{Vocabulary adaptation}\newline
Many studies in the literature focus on how to manage the vocabulary gap between domain-specific terminology and the general vocabulary of LLMs. \cite{monsin2022vocab} pointed out that traditional methods with random embeddings for expanding the vocabulary of a pretrained tokenizer have shown no gain on domain tasks compared to a DAPT model with its original tokenizer. In their study, they developed a more effective but more complex method for expanding the vocabulary of a pretrained tokenizer (VIPI) that has demonstrated strong performance. However, we have chosen not to modify the pretrained tokenizer, primarily due to the incompatibility of this method for PEFT fine-tuning techniques and the minimal improvement they offer on downstream tasks. 

\textbf{Several fine-tuning methods for DAPT}\newline
% DAPT : several ways of performing the fine-tuning
The DAPT process consists in "continuing pretraining" LLM on a domain corpus, which actually means that we are fine-tuning a LLM on the pretraining task, the same tasks on which the pretrained LLM has been initially trained.
However, with LLMs, the fine-tuning process poses significant challenges, most prominently the constraint of limited resources: as LLMs become more voluminous and complex, they are becoming progressively difficult to adapt, thus falling within the purview of a select few privileged companies with the necessary resources. 
Recently, research efforts around parameter-efficient fine-tuning PEFT~\cite{lialin2023scaling} have been geared towards developing fine-tuning methodologies that can optimally adapt these models with ``acceptable''resource expenditure, striking a balance between performance and cost. 
Thanks to those parameter efficient fine-tuning methods, DAPT can be performed in different ways: Methods like LoRA and QLoRA can help to reduce both training and inference costs. This new possibility raises the question of which method would yield the best performance for domain adaptation, and if full fine-tuning is truly superior to a more cost-effective approach or not.
\begin{enumerate}
\item Full-parameter fine-tuning is the traditional fine-tuning method, where gradient updates are applied to all the parameters of the model.
\item LoRA~\cite{hu2021lora}, which stands for Low-Rank Adaptation of LLMs, is a fine-tuning method based on the premise that the weight update matrix during fine-tuning has a larger dimension than its rank. Unlike conventional fine-tuning methods that directly modify the model's weights during backpropagation, LoRA is making these adjustments of the weights in a separate matrix. The final result of fine-tuning combines two components: the pretrained frozen model and the weight update matrix. 
The essence of the LoRA method lies in its hypotheses that the weight update matrix is, in fact, quite compact; its dimension is significantly larger than its rank. Thus, LoRA involves decomposing this weight update matrix into two rank-x matrices (e.g., 8, 16, or more). Matrix decomposition is a mathematical technique used to break down a matrix into a set of simpler component matrices. The goal is to represent a complex matrix as a combination of simpler and smaller matrices. 
The outcome of this decomposition yields what is known as an "adapter." As a result, we can load a pre-trained model off the shelf and attach its LoRA adapters (one for each fine-tuning task). Importantly, these adapters are notably compact, making them exceptionally cost-effective both in terms of storage and deployment.
\item QLoRA~\cite{dettmers2023qlora} makes some modifications to LoRA fine-tuning method: In QLoRA, the pretrained model is loaded directly as a quantized (4bit or 8bit) model (converting all its weights to 4 or 8 bits). Thus, the adapters (weight update matrix) is updated with the gradient calculated from the 4 or 8 bits pretrained model, as opposed to what we do in traditional LoRA, which is calculating the gradient from a full or half weight precision model. In QLoRA, the memory is all the more optimized that the optimizer states are stored in CPU (this process is called : “Paged Optimizers”), to optimize all the more the GPU memory. QLoRA has been implemented to allow everyone to run very large models on a single GPU. The optimizer state is all the intermediate values calculated for the backpropagation and that need to be stored to push the gradient backward.
\end{enumerate}

\textbf{DAPT + TAPT: the best combination}\newline
By comparing DAPT (fine-tuning a pretrained model on the pretraining task (= retraining) on a domain corpus) and TAPT (fine-tuning a pretrained model on the evaluation task) on 4 specialized domain (biomedical, computer science, news and reviews) on eight classification tasks,~\cite{gururangan2020domainadaptation} demonstrated that the best results are obtained with a combination of the two methods: DAPT + TAPT. This observation has been extended to larger models~\cite{Huang2023readingcomprehension}. Doing a combination of DAPT and TAPT is all the more important for LLMs as~\cite{Huang2023readingcomprehension} highlighted that for a LLM such as Llama-2-7b, DAPT-only (prompting the adapted model directly to answer questions) is degrading the original model performance on most domain tasks. Contrary to the very satisfying results of DAPT-only on Roberta-base~\cite{gururangan2020domainadaptation}, results on DAPT-only on Llama-2-7b are disappointing, even worse than a non-adapted Llama-2-7b model. DAPT seems indeed to be drastically hurting LLMs prompting ability. However,~\cite{Huang2023readingcomprehension} points out that further fine-tuning the adapted model on a specific task (QA, for instance) performs better than the original model (llama-7b) finetuned on the same task, for each task tested (such as QA, NER and sentiment analysis) and each domain (BioMed, Finance and Law).

\textbf{Identification of telco-domain downstream tasks valuable for the industry}\newline
\cite{maatouk2023large} examined different practical applications of LLMs that can be readily utilized by telecom-domain stakeholders with their existing data. One of the task they suggest is the comprehension of 3GPP specifications and the development of chatbots on these data. These chatbots, based on fine-tuned LLMs, could help engineers streamline their research, saving time and improving the accuracy of 3GPP standards implementation and related research. Another task might be the resolution of questions regarding the network modeling. LLM can be tasked to estimate energy consumption in a network based on selected features, for instance. All in all, question-answering on the network / telco legal specifications seems to be an interesting and valuable downstream task for LLMs in the telco domain.

% ----------------------------------------------
% ----------------------------------------------
% ----------------------------------------------

\section{Methodology}
\label{sec:methodology}
In this paper, we are comparing three different approaches (DAPT-only, DAPT + IAPT, IAPT-only) on Llama-2-7B~\cite{Touvron2023llama}, to find the most effective way of adapting this model to the telecommunication industry.
Since there is no consensus on the content of the instructions for domain-adaptation (domain-only instruction vs a blend of general and domain instructions)~\cite{chipnemo}, we compare the performance of the adaptation using all those combinations. 
In the end, we test 6 different adaptation methodologies : DAPT-only, DAPT-IAPT with telco instructions, DAPT-IAPT with general instructions, DAPT-IAPT with a blend of telco and general insturctions, IAPT-only with telco instructions, IAPT-only with a blend of telco and general instructions.

\subsection{Telco Dataset}
\label{sec:dataset}
Data quality plays a crucial role in producing a powerful model. 
A significant focus in recent research centers on emphasizing the importance of tasks such as data crawling, data generation and data cleaning~\cite{Penedo2023refinedweb,gunasekar2023textbooks,longpre2023pretrainer,lee2023beyond,benallal2024cosmopedia,lozhkov2024starcoder}, which lay the foundation for data-centric development of LLMs. 
Inspired by these works, our work relies on a telco-specific dataset extracted from public resources, consisting of a 803M tokens for DAPT and associated instructions for IAPT. 

\subsubsection{Pretraining data for DAPT}
\label{sec:data_dapt}
The pre-training corpus includes raw texts (803M tokens) specialized in the telco domain, from publicly online repositories (Table.~\ref{tab:dapt_data}).
% This corpus was assembled through a series of strategic crawls across various online repositories, utilizing the web scraping framework, Scrapy, to retrieve PDFs and HTML pages.
Specifically, \textit{(i)} 41\% of tokens are extracted from technical specifications, white papers published by standards organizations (28\%), such as 3GPP, ITU, ETSI, RFC, etc, as well as from research papers (13\%), published in Arxiv, in the field of Networking and Internet Architecture; \textit{(ii)} 11\% tokens are retrieved from telco-related domains in Common Crawl archives (e.g. 5GAcia, 5GAmericas, Nokia, CTIA, GSMA, Huawei, IPv6, Juniper, etc) and in Stack Exchange archives (e.g. Networking Engineering, Signal Processing, Security, Quantum Computing categories); \textit{(iii)} 47\% of tokens are selected from more diverse sources via importance sampling (Data Selection for Language Models via Importance Resampling, DSIR)~\cite{dsir}. Given the target telco dataset described in \textit{(i), (ii)}, DSIR aims to select relevant texts that are distributed like the target. With this method, we select telco-relevant samples from stack overflow (9\% tokens), wikipedia (12\%), open web mathematical texts (10\%), arXiv papers in RedPajama dataset (15\%). 

\begin{table}
\small
    \centering
    \begin{tabular}{lc}
    \hline
    \textbf{Corpus} & \textbf{Proportion}\\ \hline
    Technical Documents (3GPP, ITU, RFC, etc.) & 28\%  \\ \hline
    Arxiv & 28\% \\  \hline
    Wikipedia & 12\% \\ \hline
    Common Crawl & 11\% \\ \hline
    Open-web-math & 10\% \\ \hline
    StackOverflow & 9\%  \\ \hline
    StackExchange & 2\% \\ \hline
    \end{tabular}
    \caption{Proportion for each source}
    \label{tab:dapt_data}
\end{table}

% We applied DSIR method on StackExchange discussions (2\%), Wikipedia entries (12\%), open web mathematical texts (10\%), ArXiv papers from RedPajama (15\%) and StackOverflow Q\&A sessions (9\%), specifically filtered for telecommunications relevance. For instance, Wikipedia pages were selected with an emphasis on entities close to the Telecommunications sector, while StackExchange and StackOverflow contributions were chosen for their focus on relevant Q\&A pairs.

% To ensure the relevance and technical precision of our corpus, we employed several methods to extract the texts from those PDFs crawled. Indeed, 
% using several extraction methods allowed us to have a diversified text-format in our training corpus:
Our crawled data exists in two different formats: PDF, HTML. To make them available for use with LLM, we perform the extraction and cleansing processes, employing the following methods:
\begin{itemize}
\item Web data (HTML):  inspired by Falcon~\cite{Penedo2023refinedweb}, SlimPajama~\cite{cerebras2023slimpajama} and Gopher~\cite{rae2021scaling}, we create a holistic pipeline for refining and deduplicating web data, consisting of steps such as language detection, document-wise and line-wise filtering, exact and fuzzy deduplication.
\item PDF data: for technical and math-heavy documents, we employ Nougat~\cite{nougat}, a visual Transformer to  transform documents into markup language. For documents collected in Common Crawl,  PDFMiner is used to extract the texts. To further assess the relevance and quality of the texts extracted, the Zephyr-7b~\cite{tunstall2023zephyr} model was prompted to classify whether a text is linked to the telecommunication domain. The prompt we used for this can be found in Appendix~\ref{app:zephyrprompt}.
\end{itemize}

The validation dataset consists of 5.5M telco-specific tokens extracted from sources other than training data, and 5.5M tokens taken from general-domain corpora such as Wikipedia, PubMed and the Pile ('OpenSubtitles' and 'Enron emails' splits)~\cite{gao2020pile}. This second data split aims to acquire insights into the model's behavior in general domains as it is adapted to the Telco domain.

% \begin{figure*}[htbp]
%    \centering
%    \includegraphics[width=\linewidth]{figures/data_pipeline}
%    \caption{Dataset collection and preparation}
%    \label{fig:methodo}
% \end{figure*}

\subsubsection{Instructions for IAPT}
\label{iaptdataset}

Diverse and high-quality Telco-related instruction could enhance the instruct-tuning of Telco-adapted LLMs. Instructions for IAPT were either collected from existing material or generated. The overview of the training and validation sets for intructions is given in Table~\ref{tab:stats_instr_train_valid}. In the light of a multitude of works on using LLMs for instruction generation~\cite{wang2023self,xu2023wizardlm,OpenOrca}, we develop a partially synthetic instruction dataset, tailored to the Telco domain. The corpus is created by prompting GPT3.5 and Mixtral~\cite{jiang2024mixtral} to produce tasks instructions and their input-output instances in JSON format {\tt \{`instruction', `input', `output'\}} from pre-training raw text (Section~\ref{sec:data_dapt}), as following:
% Our instruction dataset is composed of instructions we have generated using prompt engineering (using GPT-3.5 and Mixtral models) on our DAPT corpus. The idea is to transform the DAPT training corpus into instructions. To generate IAPT corpus, GPT-3.5 and Mixtral models were provided with raw data from the corpus and generated instruction from it. 
% Here is a detailed explanation on how we have done this generation:
% \begin{itemize}
    % \item 

\begin{table}[t!]
   \centering
   \includegraphics[scale=0.8]{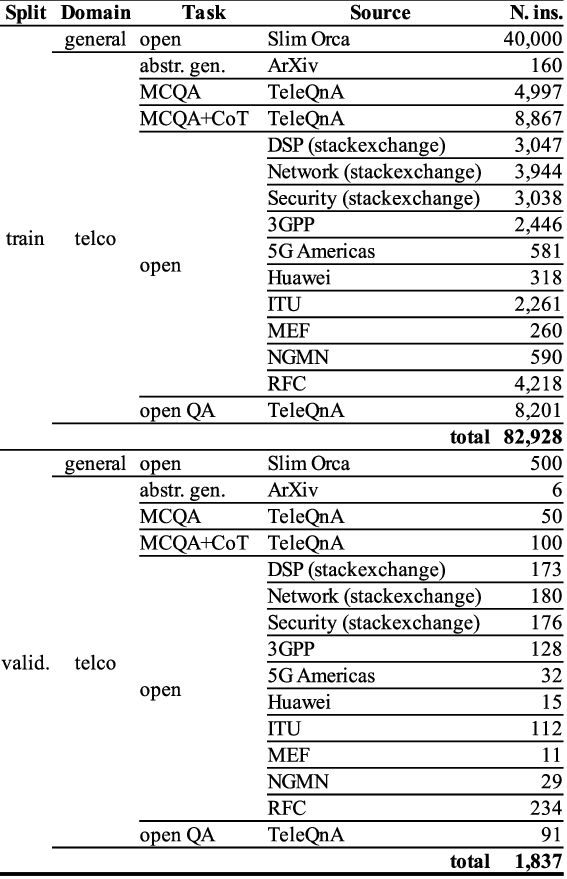}
   \caption{Statistics about training and validations instruction sets}
   \label{tab:stats_instr_train_valid}
\end{table}
    
\paragraph{Generate instructions from Telco technical documents.} 
\label{syntheticinstructions}
Considering that technical documents are too long to be entirely provided to the LLMs for instruction generation, we adopted paragraph-level generation strategy. Documents are divided into paragraphs using '\texttt{\#\#}' delimiter. Paragraphs that are less informative or exceed the context limit, such as ones containing fewer than 256 words or more than 3092 words, excessive punctuations, or insufficient alphabetic characters, are ignored. The prompt provided to the models are found in Appendix~\ref{app:promptspecsinstruction}. As a post-processing step, we eliminate generated instructions that refer to implicit information (i.e.  figure, table, section, paragraph, clause, equation, annex, paper, text, document) or those with outputs fewer than 200 characters, which are typically of lower quality. For the later, a second prompt was used to expand the output into a more detailed and informative response, as described in Appendix~\ref{app:secondaryprompt}. We obtain, with this process, a total of 11,235 instructions / output pairs. 
% The detailed composition of those instructions can be found in Table~\ref{tab:specsinstructions}. 
% You can also find in Appendix~\ref{tab:sampleinstructions1} some samples of the generated instructions.\\
    % \item 
\paragraph{Generate instructions from Telco Question-Answer pairs.} We rephrase 10,558 Stack Exchange question-answer pairs about network engineering, digital signal processing and security topics into well-formatted instructions by prompting LLMs. 
% his resulted in an additional 10,558 instructions, with a detail of this composition in Table~\ref{tab:stackexchangeinstructions}. 
% You can also find in Appendix~\ref{tab:sampleinstructions2} some samples of the generated instructions. 
In addition, we transform 4,000 multiple-choice questions from TeleQnA dataset~\cite{teleqna} into instructions in the chain-of-thought style. This is achieved by prompting LLMs to produce an explanation path leading to the correct choice for each question provided. The employed prompts are shown in Appendix~\ref{app:promptcot}.\\
% \end{itemize}

% \begin{table*}[t]
%     \centering
%     \begin{tabular}{|l|c|c|}
%     \hline
%     \textbf{Topic} & \textbf{Instruction with input} & \textbf{Instruction without input} \\ \hline
%     5GAmericas & 16 & 597 \\ \hline
%     MEF & 0 & 271 \\ \hline
%     Huawei & 4 & 329 \\ \hline
%     NGMN & 176 & 443 \\ \hline
%     3GPP & 1152 & 1422 \\ \hline
%     ITU & 1539 & 834 \\ \hline
%     RFC & 534 & 3918 \\ \hline
%     \end{tabular}
%     \caption{Instruction dataset composition for telco standards and specifications texts from DAPT corpus}
%     \label{tab:specsinstructions}
% \end{table*}

% \begin{table*}[t]
%     \centering
%     \begin{tabular}{|l|c|c|}
%     \hline
%     Topic & Instruction with input & Instruction without input \\ \hline
%     Network & 2398 & 1726 \\ \hline
%     DSP & 2096 & 1124 \\ \hline
%     Security & 2001 & 1213 \\ \hline
%     \end{tabular}
%     \caption{Instruction dataset composition for telco Stack Exchange Discussions from DAPT corpus}
%     \label{tab:stackexchangeinstructions}
% \end{table*}

Along with Telco instructions, we have also compiled a dataset of general instruction sets. We used the Slim Orca general instructions from Open-Orca~\cite{orca}. This dataset serves to complement the specialized instructions by introducing a diverse range of general tasks.

In order to transform MCQs into instructions into our IAPT training set, we added a prefix to the question-option-choices MCQs. For general datasets, we added only the instruction "\texttt{Which of a, b, c, and d is the right answer to the following question? {question} {a. choice1, b. choice2, c. choice3, d. choice4}}" (adapting it to the amount of choices), while for telco MCQs datasets, we added the instruction "\texttt{\textbf{You are a helpful assistant, specialized in telecommunication technologies.} Which of a, b, c, and d is the right answer to the following question? \{question\} \{a. choice1, b. choice2, c. choice3, d. choice4\}}". This transformation allowed us to use these MCQs as instructions in our IAPT training set.

%TODO Slides 16 à 30 de la présentation GTC_Orange_final.pptx
%TODO Afficher clairement The OpenTelco Dataset comme contribution du papier + lien vers dépôt open-source

The development of a comprehensive benchmark for evaluating the knowledge of different models in the telecommunications sector represents a significant step forward in the domain of artificial intelligence research. Our benchmark encompasses perplexity measurement on raw texts, as well as task resolution. In each case, datasets from the telco and general domain were used. Studied tasks are Multiple-Choice Question Answering (MCQA), Open Question Answering (Open QA) and abstract generation from scientific papers. These tasks were meticulously designed to assess the models' understanding of telecommunications-related information. Statistics of the test datasets are given in Tables~\ref{tab:stats_text_test} and~\ref{tab:stats_instr_test}, for texts and task intructions respectively.

\subsubsection{Evaluation datasets}
\label{benchmarks}

\begin{table}[t!]
   \centering
   \includegraphics[scale=0.8]{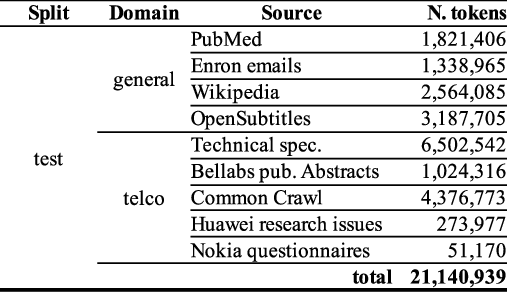}
   \caption{Test set of texts}
   \label{tab:stats_text_test}
\end{table}

\begin{table}[t!]
   \centering
   \includegraphics[scale=0.8]{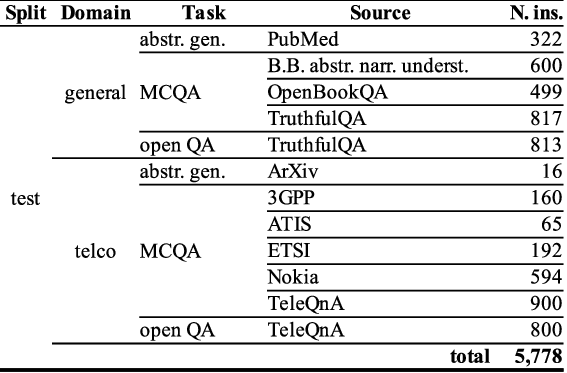}
   \caption{Test set of instructions}
   \label{tab:stats_instr_test}
\end{table}

For the MCQA telco benchmark, we have generated synthetic MCQs from select sources within our DAPT training set, namely ATIS, 3GPP, and ETSI standards, using GPT-4 with a 32k token model. To do so, we have selected some paragraphs having fewer than 256 words or more than 3092 words (as described in \ref{syntheticinstructions}), and prompted GPT4-32k to generate MCQs questions based on each paragraphs. The prompt used for this generation can be found in Appendix~\ref{app:mcqprompt}. You can also find some samples from the generated instruction test set in Appendix~\ref{app:daptinstructions}. Each MCQ generated has a single correct answer and 5 choices per question. For this type of MCQs, questions are about knowledges present in the DAPT training set, since we have generated these MCQs based on these knowledges. Additionally, we incorporated to our telco MCQ benchmark the TeleQnA test set from Huawei \cite{teleqna}, composed of 900 questions. The original TeleQnA dataset was composed of 10k instructions that were split into train and test sets (9100/900 questions) as we described in \ref{iaptdataset}.  The TeleQnA dataset is a MCQ dataset referenced as the first benchmark dataset designed to evaluate the knowledge of Large Language Models (LLMs) in telecommunications. Huawei generated this dataset using diverse sources, including telecom standards and research articles. Each question has 4 to 5 choices and several potential correct answers. For this type of MCQs, we consider that some questions might be regarding knowledges present in the DAPT training set since TeleQnA was constructed based on the same sources. Finally, we have added to our telco MCQ benchmark real-world MCQs we extracted from signal processing exam questions of Nokia's official practice exams. We have extracted 632 MCQs (questions, choices and answers) from Nokia official practice exams for the Nokia NRS I, NRS II and SRA written exams. These MCQs are composed of 4 to 5 choices and multiple potential correct answers per question. For this type of MCQs, we consider that questions are far from the knowledges present in the DAPT training set since NOKIA MCQs are about signal processing questions, which are not exactly the topics we have in the DAPT training set. So these MCQs' answers are not explicitly present in any text or instruction from domain adaptation.

As for the QA telco benchmark, it is exclusively composed of the TeleQnA \cite{teleqna} test set from Huawei, composed of 900 questions (the same we used for MCQs), but here, we have converted each MCQ into questions – answers pairs.

Furthermore, to ascertain the impact of domain adaptation on retaining general knowledge, we introduced a general benchmarks. As for the general MCQ benchmark, it includes a variety of MCQs from the OpenBookQA (test set of OpenBookQA official MCQs dataset~\cite{openbookqa2018} (500 MCQs) consisting of multiple-choice elementary-level science questions with a single good answer per question and 4 choices), TruthfulQA (the subset "multiple choice" of TruthfulQA official MCQs dataset~\cite{truthfulqa2021} using the MC1 target, composed of 817 MCQs on general knowledges with a single good answer per question and 4 choices), and Big-bench datasets (subset "abstract narrative understanding" of the validation set of Big-bench dataset~\cite{bigbench2022}, composed of 600 MCQs on questions about narrative understanding of a given text, with a single good answer per question and 5 choices), encompassing elementary-level science questions, general knowledge queries, and narrative understanding assessments. As for the general QA benchmark, it is composed of the "generation" subset of the TruthfulQA dataset~\cite{truthfulqa2021}, which comprises 817 examples within its validation set. The questions are designed to evaluate the model's ability to generate coherent, accurate, and truthful answers across a spectrum of topics, thereby testing the model's grasp of factual information.

\subsection{Adaptation strategies}

\subsubsection{DAPT}

% \subsubsection{DAPT training process}
We preprocessed the dataset by tokenizing the text and segmenting it into chunks of 4096 tokens each. We used the pretrained 'meta-llama/Llama-2-7b-hf' model and tokenizer. To harness the computational power necessary of the training, we deployed the training across 2 A100 80GB GPUs, utilizing data parallelism to effectively manage and distribute the computational workload.
We adhered to a batch size was set to 32, using micro-batch size of 1 (gradient accumulation). This method effectively simulates a larger batch size of 32, enabling us to maintain the stability and benefits of larger batch training dynamics without the associated hardware demands. The training used a learning rate of~$2\times10^{-5}$ with a cosine scheduler.
We loaded the model in half-precision (bfloat16) to balance computational efficiency with model performance. Additionally, we applied a weight decay of 0.01 to regularize and prevent overfitting. The entire training process was completed over the span of a single epoch, ensuring that the model was exposed to the full range of data while preventing the potential for overfitting that multiple epochs might introduce.

\subsubsection{IAPT}
%Instruction-Adaptive Pretraining (IAPT) consists in continuing the pretraining process on data structured into instruction-output pairs. While DAPT focuses on acclimatizing the model to domain-specific language and information, IAPT takes this a step further by structuring the pretraining process around instructions. This method involves self-supervised fine-tuning on a dataset formatted explicitly as a series of instructions, aiming to enhance the model's ability to follow commands and generate outputs accordingly.

%In IAPT, the dataset is composed of blocks of 4096 tokens, each of which consists of an instruction and its corresponding output. These blocks are marked with special tokens [INST] and [/INST] to delineate the instruction from the output. During the pretraining auto-regressive phase, the model performs backpropagation on the loss from all tokens (both the instruction and the output), which means it learns from the instruction and their answers. This enables the model to not only understand domain-specific content but also to act upon instructions related to that content.

%Implementing IAPT is akin to performing DAPT but with an additional preprocessing of the data due to the instruction format. To prepare for IAPT, one must first generate a set of instructions from the domain-specific raw dataset used in DAPT. This process involves creating prompts that guide a language model to transform domain-specific information into actionable instructions paired with appropriate outputs. 

% \subsubsection{IAPT training process}
In the IAPT process, instructions were formatted with special tokens, denoting the start \texttt{[INST]} and end \texttt{[/INST]} of each instruction, followed by the expected output. We then tokenize the instructions, and then concatenate them into blocks of 4096 tokens, ensuring that no instruction was cut off between blocks by using padding at the end of the blocks as necessary. Then, we fine-tuned the model on the auto-regressive training process.
we employed the Deepspeed Zero3 parallelization of the training, which is an advanced system designed to make the parallelization over not only the data and the optimizer states, but also the weights of the model~\cite{deepspeed}. We use here 2~NVIDIA A100 GPUs (40GB each).
We use for hyperparameters a learning rate of~$2\times10^{-5}$ with a cosine scheduler and a weight decay of 0.1 to regularize the training and prevent overfitting.
We train the model with a batch of size 64 with micro-batch of size 1. The training was carried out over 2 epochs.

We launched several IAPT trainings to find the most efficient way of adapting LLM to the telco domain: The IAPT was launched across various combinations of data, incorporating both domain-specific telco instructions and a set of general instructions to provide a comprehensive learning scope for the models.

% ----------------------------------------------
% ----------------------------------------------
% ----------------------------------------------

\section{Evaluations}
\label{sec:evaluation}

\begin{table*}[t!]
\centering
\includegraphics[scale=0.75]{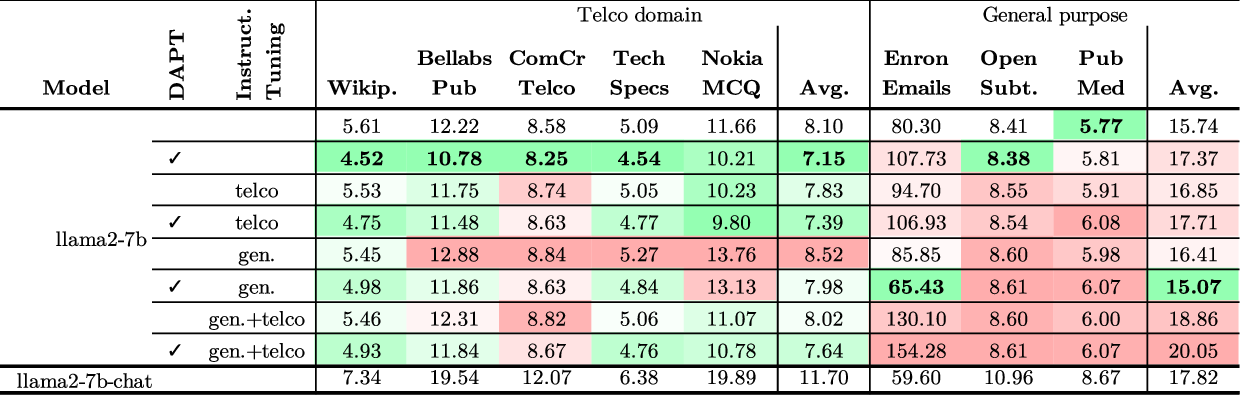}
\caption{Perplexities for each dataset and model. The best metrics per dataset is in bold and underlined, and the 2nd best metric per dataset is in bold.}
\label{tab:perplexity}
\end{table*}

\subsection{Intrinsic evaluation: perplexity}
%Perplexity is defined as the exponentiated average negative log-likelihood of a test set. It is a measurement of how well a probability distribution or probability model predicts a sample. Here, we take the test set from the DAPT raw text corpus. We take the average perplexity for each batch and each text a corpus from the test set. By taking the average perplexity per corpus, we get a single metric that represents the average uncertainty across all texts within that corpus. This average provides a concise and clear indication of the model's predictive power: the lower the average perplexity, the more effectively the model has adapted to the language characteristics of the domain-specific corpus in question. 
Perplexity measures a model's predictive ability by averaging its performance across a test corpus, with lower values indicating better adaptation to the corpus's language traits. Perplexity comparisons require models to use the same tokenizer because the tokenizer directly influences the size and composition of the model's vocabulary, which are critical factors in calculating perplexity. Consequently, we limited our comparisons to models of the LLaMA-2 type to ensure consistency in tokenizer usage and accurate perplexity evaluations. The perplexity of each adapted models can be found in Table~\ref{tab:perplexity}. 

As anticipated, the DAPT-only model demonstrates the best perplexity results on DAPT test set. Interestingly, the IAPT models, when applied to telco-only instructions, achieves perplexity metrics that are quite comparable to the DAPT-only model. This suggests that the IAPT model, despite not being solely focused on the telco domain, is effective at adapting to domain-specific language.

\subsection{MCQs results}

In evaluating model performance, we primarily use multiple-choice questionnaires (MCQs) because evaluation on multiple-choice questions is a standard practice and is generally considered within the community to be a reliable proxy for testing a model's knowledge. This conviction stems from the straightforward, binary nature of MCQs, where each question presents a set of options, and only one holds the key to accuracy. Consequently, we employ 'accuracy' as our metric of choice. This binary framework, underpinning our MCQs, allows for an unequivocal evaluation, as the model's response is distilled to a singular letter, leaving no room for ambiguity—only the precision of 'correct' or the finality of 'incorrect'.

For inference on the datasets, the Llama-2-chat instruction template (\texttt{[INST]} \{instruction\} \texttt{[/INST]}) was consistently used for all Chat and IAPT models inferences. As for GPT3.5, GPT4, Base models and those with DAPT-only, the instruction was directly fed to the model (without the \texttt{[INST]} and \texttt{[/INST]} tags). %Instructions were tailored to match the dataset context: for telco datasets with 4 choices, the instruction was "You are a helpful assistant, specialized in telecommunication technologies. Which of a, b, c, and d is the right answer to the following question? \{question\} \{a. choice1, b. choice2, c. choice3, d. choice4\}", while for general datasets, the instruction was "Which of a, b, c, and d is the right answer to the following question? {question} {a. choice1, b. choice2, c. choice3, d. choice4}". Instructions varied based on the number of options, adapting for MCQs with different choice counts, including those with 5 options.

We launched the inference on the different Llama-2-7b-dapt, Llama-2-7b-dapt-iapt and Llama-2-7b-iapt models. As baseline commparison, we also launched the inference on meta-llama/Llama-2-7b-hf, on meta-llama/Llama-2-7b-chat-hf, as well as on GPT3.5 and GPT4. Note that the chat model from Huggingface (meta-llama/Llama-2-7b-chat-hf) is trained, in addition to instruction-tuning on meta-llama/Llama-2-7b-hf, with RLHF-based training, which differs from our Llama-2-7b-iapt and Llama-2-7b-dapt-iapt models. So our instruct-tuned models and the chat model from Llama-2 are not directly comparable. %Thus, these models are not really comparable. A better comparison would be between the iapt-telco+general or iapt-telco-only models and the iapt-general-only model, which is alike the Llama-2-7b-chat-hf model without RLHF.

We've set up a detailed process to clean up the answers the different models give. First, we use regular expressions (regex) to pick out the answer letters from different types of responses, like 'a. choice' or 'the answer is a.'. If the answers are not clear enough for this method, we use GPT-3.5 to retrieve the answer letters from natural language responses.

\begin{table*}[t]
    \centering
    \includegraphics[scale=0.75]{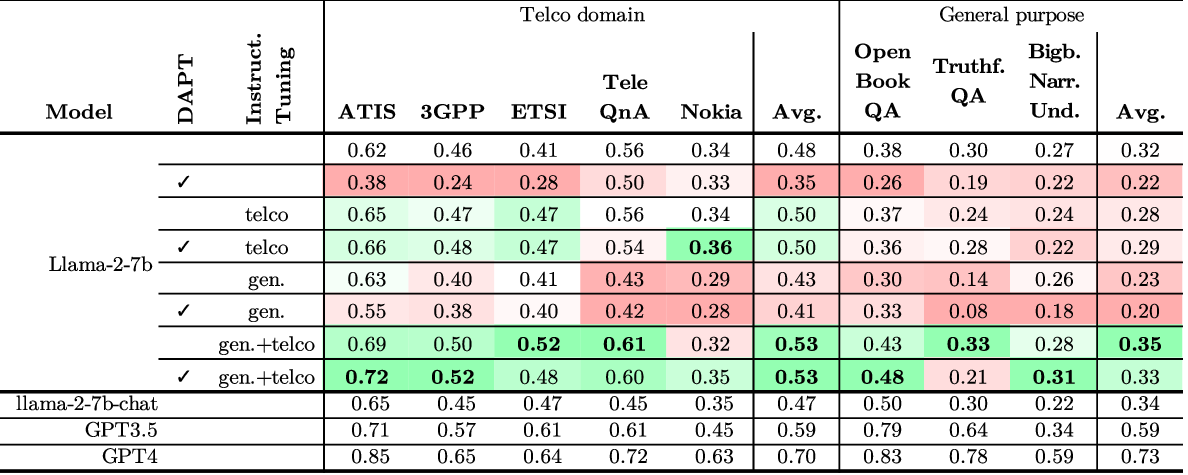}
	\caption{Comparison of models accuracy on different types of MCQs. Color shades are compared to the performance of the base model (i.e., LLaMA-2 before adaptation). Red and green colors mean that the results are worse or better, respectively. Bold scores are for the best setting among the adapted LMs.}
	\label{tab:mcqs}
\end{table*}

As exposed in Table~\ref{tab:mcqs}, evaluations were performed on telco MCQs and general MCQs.

\subsubsection{MCQs results on telco MCQs}
Regarding the telecommunications-specific MCQs benchmark, as described in \ref{benchmarks}, we have 3 types of datasets:
\begin{itemize}
\item MCQs with questions related to knowledge present in the DAPT and with similar IAPT instructions (ATIS, 3GPP, ETSI)
\item MCQs with some questions which might be related to knowledge present in the DAPT and with similar IAPT instructions (TeleQnA)
\item MCQs whose answer is not explicitly present in any text or instruction from domain adaptation (Nokia)
\end{itemize}

Here are our observations:
\begin{itemize}
\item We first observe the same conclusion as~\cite{readingcomprehension2023} that DAPT-only adaptation results in a drastic drop in prompting performance of the model, making it unusable for domain-specific downstream tasks. However, coupled with IAPT adaptation, DAPT seems to result in telco-MCQ performance improvement. But the gain of DAPT (compared to IAPT-only adaptation) is not obvious (accuracy improved by +0.03 maximum). This leads us to conclude that DAPT is not necessary for domain adaptation, and that IAPT-only is sufficient.
\item We then observe that IAPT incorporating both telco and general instructions proves most effective, enhancing accuracy by +0.09 on average compared to the chat model for similar task datasets. However, telco-only IAPT modestly boosts outcomes by +0.03 and IAPT focusing solely on general instructions slightly decreases telco MCQs performance by -0.04. 
\item We finally observe that our methods (dapt+iapt or iapt-only) are especially beneficial for specific knowledge MCQs present either in the DAPT corpus or in the IAPT one, or in both. The results on the Nokia dataset were notably less favorable, primarily because this corpus demands a broader foundational knowledge than what is provided by the technical documents and instructions utilized in the adaptation process. This discrepancy highlights a crucial aspect of domain adaptation: while it significantly enhances a model's proficiency within a specific domain, its capacity for generalization beyond the scope of the adapted materials is inherently limited.
%Indeed, MCQ task on topics beyond DAPT knowledge, such as Nokia's signal processing MCQs, does not exhibit improvements compared to the Chat model with our domain adaptation method, highlighting this method's limitation in adapting to inter-domain knowledge.
\end{itemize}

\subsubsection{MCQs results on general MCQs}
We have also conducted our experiments on general datasets to analyze whether or not domain adaptation leads the model to forget general knowledge. The general MCQs benchmarks is described in \ref{benchmarks}.

Results of our experiences on these datasets can be found in Table~\ref{tab:mcqs}. Here are our observations:

\begin{itemize}
\item As expected, the different adapted models do not outperform Llama-2-7b-chat-hf on General Knowledge MCQs. 
\item We do not observe a too-big catastrophic forgetting with our methodolody. Indeed, the adapted models have a decrease in accuracy of -0.02 with the DAPT-IAPT on general + telco instructions: it's even noteworthy that our DAPT+IAPT method, utilizing both general and telco instructions, slightly boosts performance (+0.03 in accuracy) compared to the chat model on the TruthfulQA dataset. 
\item Interestingly, the Llama-2-7b-IAPT and Llama-2-7b-DAPT-IAPT models, when solely based on general instructions, underperforms, possibly due to the poor quality or insufficient quantity of general instructions (40K instructions from SlimOrca~\cite{orca}) in the IAPT train set as the source for these general instructions).
\item On text understanding MCQs, it appears our models perform competently, matching or even surpassing the baseline models. This suggests a solid grasp of broader knowledge and comprehension. This suggests that catastrophic forgetting tends to be more pronounced in certain general knowledge domains than in comprehension tasks.
\end{itemize}

\subsection{Open QA results}
We employ a context-free question answering (QA) approach as part of our test set to assess our models. This particular methodology, known as context-free QA, involves posing questions that do not rely on a given text or background information. 
For evaluating the performance of our models on this context-free QA setup, we utilize the ROUGE and METEOR metrics. ROUGE, which stands for Recall-Oriented Understudy for Gisting Evaluation, primarily focuses on how many of the same words and phrases appear in both the model's response and the reference answer, essentially measuring the overlap and thereby the accuracy of content. METEOR, on the other hand, stands for Metric for Evaluation of Translation with Explicit Ordering. It goes a step further by not only accounting for the similarity in terms of word overlap but also considering synonyms and the order of the words, providing a more nuanced assessment of the model's linguistic precision and fluency. Both metrics together offer a comprehensive view of the model's capability in generating accurate and contextually appropriate answers.
However, this observation is limited by the ROUGE and METEOR metrics. Despite being useful for evaluating text generation models, these metrics have notable limitations: ROUGE primarily measures n-gram overlap, which might not capture the deeper semantic meaning or the coherency of the text, potentially leading to high scores for texts that are lexically similar but contextually different. METEOR, while incorporating synonyms and basic paraphrasing, still falls short in fully grasping nuanced language usage and does not effectively evaluate the logical structure or factual accuracy of longer content. 

To overcome the limitations of those traditional metrics, we added an LLM scoring approach, leveraging GPT4-128k as the scorer. Following the methodology outlined in the paper Prometheus \cite{kim2023prometheus}, we provided GPT4 with both the correct reference answer and the model's generated answer for each question and each model prediction. We then prompted the model to assign a score from 1 to 5 to each response. As described in the Prometheus paper, we provided detailed guidance on the interpretation of each score within the prompt sent to GPT4, with additional specifics outlined in the Appendix~\ref{app:promptllmscore}.

\subsubsection{Open QA results on telco}
Our dataset for this task is derived from Huawei's TeleQnA dataset (as we used in MCQ task): we adapted this TeleQnA multiple-choice question dataset into a context-free question and answer dataset, streamlining the data to focus purely on question-driven responses. In this transformation, we retained only the question component from the original dataset, omitting any contextual or multiple-choice elements. To guide the model towards concise responses, we appended the instruction "Answer briefly (in a few words)." to each line. In this setup, the model operates without contextual inputs, relying solely on the question to generate its predictions. The absence of context places the emphasis on the model's inherent knowledge and understanding, challenging it to deliver precise and relevant answers based solely on the question presented.

Results of our experiences on these datasets can be found in Table~\ref{tab:qawithoutcontext}. Here are our observations:
\begin{itemize}
\item It seems that dapt-iapt-telco + general, dapt-iapt-telco-only and iapt-telco + general models tend to generate answers more similar to the expected ones.
\item It seems that dapt-iapt-telco+general and dapt-iapt-telco-only, as well as iapt-telco+general and iapt-telco-only tend to produce more concise responses than the other models, being closer to the length of the expected answers (6.79 words per question in average).
\end{itemize}

\begin{table}[t]
\centering
\includegraphics[scale=0.75]{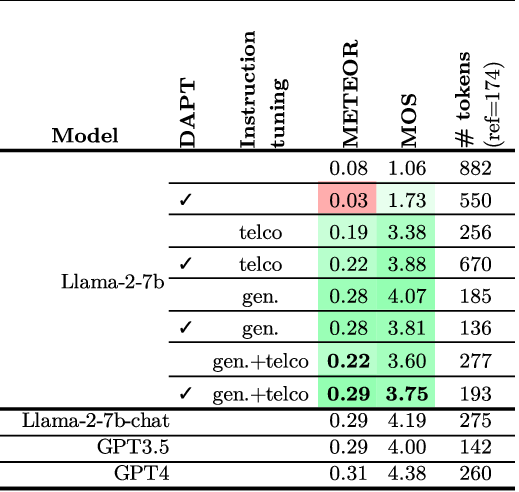}
\caption{Comparison of model's performance metrics on telco open QA. The dataset is the test set of TeleQnA transformed into open QA on questions with single answers.  The real answers' average length is 6.79 words. Color shades are compared to the performance of the base model (i.e., LLaMA-2 before adaptation). Red and green colors mean that the results are worse or better, respectively. Bold scores are for the best setting among the adapted LMs.}
\label{tab:qawithoutcontext}
\end{table}

These observations seems to be confirmed when looking at the answers: detailed responses to a selection of questions from the Open QA task are provided in the Appendix~\hyperref[app:samplestelcoqawithoutcontext]{Comparison of models answers for telco Open QA task}. We observe indeed that the models dapt-iapt-telco+general and dapt-iapt-telco-only, as well as iapt-telco+general and iapt-telco-only, tend to produce more concise responses that directly address the query. In contrast, models such as GPT-4, GPT-3.5, meta-llama/Llama-2-7b-chat-hf, and the dapt-iapt-general-only and iapt-general-only, typically generate significantly lengthier responses.

Furthermore, when reviewing the answers across numerous questions, it becomes evident that only the dapt-iapt-telco+general, dapt-iapt-telco-only and iapt-telco+general models consistently provide correct responses. This trend is particularly noticeable in the results presented in Tables~\ref{tab:sample1},~\ref{tab:sample2}, and~\ref{tab:sample3} of Appendix~\ref{app:samplestelcoqawithoutcontext}. Among these, the dapt-iapt-telco+general model stands out, uniquely delivering accurate answers in the contexts of Table~\ref{tab:sample4} and Table~\ref{tab:sample6}, indicating its superior performance in this specific evaluation.

\subsubsection{Open QA results on general domain}
We are focusing here on the "generation" subset of the TruthfulQA dataset, which comprises 817 examples within its validation set. The questions are designed to evaluate the model's ability to generate coherent, accurate, and truthful answers across a spectrum of topics, thereby testing the model's grasp of factual information.

\begin{table}[t]
\centering
\includegraphics[scale=0.75]{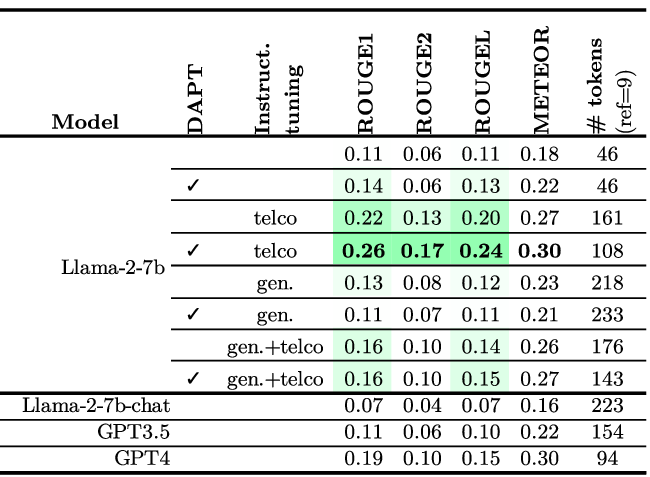}
\caption{Comparison of model's performance metrics on TruthfulQA for the general domain open QA task. The real answers' average length is 9.22 words. Color shades are compared to the performance of the base model (i.e., LLaMA-2 before adaptation). Red and green colors mean that the results are worse or better, respectively. Bold scores are for the best setting among the adapted LMs.}
\label{tab:generalqawithoutcontext}
\end{table}

Results of our experiences on these datasets can be found in Table~\ref{tab:generalqawithoutcontext}. Here, the ROUGE and METEOR metrics do not appear to be particularly relevant as they seem to favor shorter responses, especially for the "dapt-iapt-telco-only" and "iapt-telco-only" models, particularly in the case of the ROUGE metrics. However, when we delve into the detailed results (see Appendix~\ref{app:samplesgeneralqawithoutcontext}, each model give a similar answer. We do not observe any degradation of the results with our adaptation method. However, this might be due to the nature of the questions of this dataset, which  tend to be more on the intuitive side, and the answers can sometimes be open to multiple interpretations, as seen in questions like "How many days did it take to create the world?". Here, we encounter a similar pattern as with multiple-choice general questions, where for questions that do not require deep general knowledge, our domain adaptation method does not seem to degrade the results.

\subsection{Abstract generation results}

The last evaluation task is abstract generation. This tasks is less focused on directly measuring the telco knowledge injected in the adapted LMs but rather to see if this knowledge can help in a more complex task. Based on an input arXiv paper from which the text of the body has been extracted, the task is about generating the abstract that could correspond to it. This is slightly different from a standard summarization task since abstract usually follow a given structure (e.g., context, focus, experiments and key results). Furthermore, this is an intereting evaluation setup because, contrary to QA, very few examples have been shown for this task to the adapted LMs during the adaptation. To avoid token limit errors, only paper with input text shorter than 3K tokens are considered. The papers are from arXiv in the domain of network and telecommunications.

\begin{table}[t]
\centering
\includegraphics[scale=0.75]{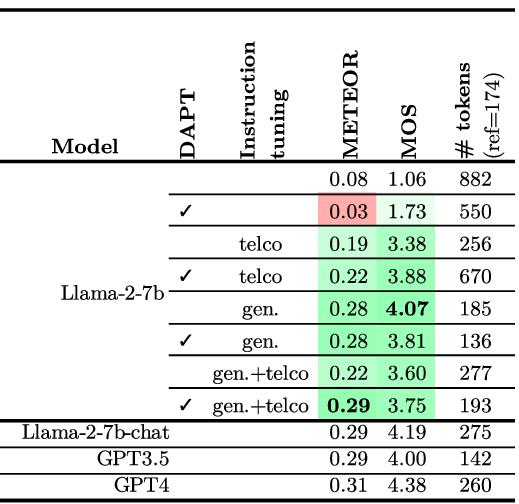}
\caption{Results of the models on the telco abstract generation task. Color shades are compared to the performance of the base model (i.e., LLaMA-2 before adaptation). Red and green colors mean that the results are worse or better, respectively. Bold scores are for the best setting among the adapted LMs.}
\label{tab:abstract_generation}
\end{table}

The results are given in Table~\ref{tab:abstract_generation}. The quality of each abstract is measured against the reference abstract in terms of METEOR and Mean Opinion Score (MOS, between 1 and 5) computed by GPT4-128k. For a better analysis, the length of each generated abstract is given. The mean length of reference abstracts is $174$. The results show that adaptation clearly leads to improvement over the baseline Llama-2-7b model. However, the benefits seems to come from the IAPT step, whatever the set of instructions is used. This tends to conclude that learning the instruction format to the model is enough. Indeed, while the best METEOR score is when using telco+general instructions, the best most MOS is when using only general instructions. This raises questions about the need for more diversity in the instructions to leverage knowledge in diverse tasks.

% ----------------------------------------------
% ----------------------------------------------
% ----------------------------------------------

\section{Conclusion and Future Work}
\label{sec:conclusion}

In this paper, we have shown a pipeline to build TelcoLMs and evaluate them. Based on Llama-2, we have conducted continued pretraining and instruction-tuning of collected texts and instruction for the telecommunication domain.
Our findings challenge the prevailing notion that DAPT, which centers on the continuous pretraining using raw, domain-specific texts, is essential for enhanced model performance within the telecommunications sector. Instead, our research indicates that IAPT, which involves ongoing pretraining on raw texts from the domain but structured as instructions, yields more substantial improvements across a variety of downstream tasks in telecommunications. Also, it seems that IAPT is sufficient (very few further gain when combined with DAPT).
We have also discovered that the most effective application of IAPT occurs when domain-specific instructions are blended with general instructions, similar to the Orca dataset. This hybrid approach leverages the nuanced knowledge of the domain while enhancing the model's capacity to generalize across a broader spectrum of tasks.
We found that the construction of the training corpus is paramount, as it must closely encapsulate the domain's core knowledge. It appears that such domain-specific adaptation methods do not inherently equip the model with the flexibility to effectively tackle cross-domain knowledge — diverging slightly from the data used in the adaptation phase. This suggests a potential limitation in the domain-adaptive approach when dealing with tasks that require broader knowledge.

Further investigation is warranted to determine the true extent of the benefits offered by IAPT within the telecommunications sector. It would be prudent to explore a broader range of domain-specific use cases to fully understand the potential advantages of this adaptation method. Such explorations would inform whether the improvements observed are consistently applicable across various tasks within the domain or if they are confined to a subset of tasks that are closely aligned with the structured training data.

\bibliography{main.bib}

\onecolumn

\appendix

\section{Comparison of models answers for telco Open QA task}
\addcontentsline{toc}{section}{Comparison of models answers for telco Open QA task}
\label{app:samplestelcoqawithoutcontext}

\begin{table*}[h!]
\footnotesize
\centering
\begin{tabular}{p{2.7cm}p{12.5cm}}
\hline
\multicolumn{2}{c}{\textbf{INPUT}} \\ \hline
\textbf{System} & You are specialized in Telecommunication domain. You are familiar with topics like 5G, RAN, Wireless communication, etc as well as technical Telco standards, specifications such as 3GPP, ITU, RFC, etc. \\ \hline
\textbf{Instruction} & Which virtualization technology enables the creation of multiple network instances optimized for specific services? Answer briefly (in a few words). \\ \hline
\multicolumn{2}{c}{} \\ 
\multicolumn{2}{c}{\textbf{REFERENCE}} \\ \hline
\textbf{Real Answer} & Network Slicing \\ \hline
\multicolumn{2}{c}{} \\ 
\multicolumn{2}{c}{\textbf{MODEL PREDICTIONS}} \\ \hline
\textbf{gpt4} & Network Function Virtualization (NFV) \\ \hline
\textbf{gpt3.5} & Network Function Virtualization (NFV) enables the creation of multiple network instances optimized for specific services. \\ \hline
\textbf{Llama-2-7b-hf-dapt-iapt-telco+general} & Network slicing \\ \hline
\textbf{Llama-2-7b-hf-iapt-telco+general} & Network slicing \\ \hline
\textbf{Llama-2-7b-hf-dapt-iapt-general-only} & The virtualization technology that enables the creation of multiple network instances optimized for specific services is called Network Function Virtualization (NFV). \\ \hline
\textbf{Llama-2-7b-hf-iapt-general-only} & The virtualization technology that enables the creation of multiple network instances optimized for specific services is called network function virtualization (NFV). \\ \hline
\textbf{Llama-2-7b-hf-dapt-iapt-telco-only} & Network slicing \\ \hline
\textbf{Llama-2-7b-hf-iapt-telco-only} & Network Slicing \\ \hline
\textbf{Llama-2-7b-hf-dapt} & A: The answer is: Virtualization technology enables the creation of multiple network instances. The virtualization technology enables the creation of multiple network instances. \\ \hline
\textbf{Llama-2-7b-chat-hf (not modified)} & Answer: SDN (Software-Defined Networking) \\ \hline
\textbf{Llama-2-7b-hf (not modified)} & \#\#\# Explanation - Nested Virtualization (https://en.wikipedia.org/wiki/Nested\_virtualization) \#\#\# References - Nested Virtualization (http://www.vmware.com/resources/techresources/1007/nested-virtualization.pdf) \\ \hline
\end{tabular}
\caption{Example 1}
\label{tab:sample1}
\end{table*}

\begin{table*}[h!]
\footnotesize
\centering
\begin{tabular}{p{2.7cm}p{12.5cm}}
\hline
\multicolumn{2}{c}{\textbf{INPUT}} \\ \hline
\textbf{System} & You are specialized in Telecommunication domain. You are familiar with topics like 5G, RAN, Wireless communication, etc as well as technical Telco standards, specifications such as 3GPP, ITU, RFC, etc. \\ \hline
\textbf{Instruction} & What is one of the main functions provided by the network layer in IoT (Internet of Things)? Answer briefly (in a few words). \\ \hline
\multicolumn{2}{c}{} \\
\multicolumn{2}{c}{\textbf{REFERENCE}} \\ \hline
\textbf{Real Answer} & Routing \\ \hline
\multicolumn{2}{c}{} \\
\multicolumn{2}{c}{\textbf{MODEL PREDICTIONS}} \\ \hline
\textbf{gpt4} & One of the main functions provided by the network layer in IoT is routing data packets from the source to the destination. \\ \hline
\textbf{gpt3.5} & One of the main functions provided by the network layer in IoT is routing data packets between devices and networks. \\ \hline
\textbf{Llama-2-7b-hf-dapt-iapt-telco+general} & Routing of data packets \\ \hline
\textbf{Llama-2-7b-hf-iapt-telco+general} & Routing of packets \\ \hline
\textbf{Llama-2-7b-hf-dapt-iapt-general-only} & One of the main functions provided by the network layer in IoT (Internet of Things) is to ensure communication between devices and the internet. \\ \hline
\textbf{Llama-2-7b-hf-iapt-general-only} & The network layer in IoT (Internet of Things) is responsible for providing end-to-end connectivity between devices and the internet, ensuring that data is transmitted and received efficiently. \\ \hline
\textbf{Llama-2-7b-hf-dapt-iapt-telco-only} & To provide a reliable and efficient data transmission service \\ \hline
\textbf{Llama-2-7b-hf-iapt-telco-only} & Routing \\ \hline
\textbf{Llama-2-7b-hf-dapt} & Provides a common addressing scheme for all devices in the network. \\ \hline
\textbf{Llama-2-7b-chat-hf (not modified)} & One of the main functions of the network layer in IoT is to provide interoperability among different devices and systems, enabling them to communicate and exchange data seamlessly. \\ \hline
\textbf{Llama-2-7b-hf (not modified)} & The network layer is responsible for routing data packets from one device to another. This includes determining the best route for the data to take, as well as managing the flow of data between devices. \\ \hline
\end{tabular}
\caption{Example 2}
\label{tab:sample2}
\end{table*}

\begin{table*}[h!]
\footnotesize
\centering
\begin{tabular}{p{2.7cm}p{12.5cm}}
\hline
\multicolumn{2}{c}{\textbf{INPUT}} \\ \hline
\textbf{System} & You are specialized in Telecommunication domain. You are familiar with topics like 5G, RAN, Wireless communication, etc as well as technical Telco standards, specifications such as 3GPP, ITU, RFC, etc. \\ \hline
\textbf{Instruction} & Which layer in the OSI (open systems interconnection) protocol architecture is designed for delivering packets from a source node to a destination node through intermediate routers based on their IP (Internet protocol) addresses? Answer briefly (in a few words). \\ \hline
\multicolumn{2}{c}{} \\
\multicolumn{2}{c}{\textbf{REFERENCE}} \\ \hline
\textbf{Real Answer} & Network layer \\ \hline
\multicolumn{2}{c}{} \\
\multicolumn{2}{c}{\textbf{MODEL PREDICTIONS}} \\ \hline
\textbf{gpt4} & The Network Layer (Layer 3) \\ \hline
\textbf{gpt3.5} & The Network layer (Layer 3) in the OSI protocol architecture. \\ \hline
\textbf{Llama-2-7b-hf-dapt-iapt-telco+general} & Network layer \\ \hline
\textbf{Llama-2-7b-hf-iapt-telco+general} & Network layer \\ \hline
\textbf{Llama-2-7b-hf-dapt-iapt-general-only} & The "Transport" layer in the OSI protocol architecture. \\ \hline
\textbf{Llama-2-7b-hf-iapt-general-only} & The Transport layer in the OSI protocol architecture. \\ \hline
\textbf{Llama-2-7b-hf-dapt-iapt-telco-only} & Network layer \\ \hline
\textbf{Llama-2-7b-hf-iapt-telco-only} & Data-link layer \\ \hline
\textbf{Llama-2-7b-hf-dapt} & OSI protocol architecture layer designed for delivering packets based on their IP addresses. \\ \hline
\textbf{Llama-2-7b-chat-hf (not modified)} & The Data Link Layer in the OSI protocol architecture. \\ \hline
\textbf{Llama-2-7b-hf (not modified)} & The transport layer is responsible for delivering packets based on their IP addresses. \\ \hline
\end{tabular}
\caption{Example 3}
\label{tab:sample3}
\end{table*}

\begin{table*}[h!]
\footnotesize
\centering
\begin{tabular}{p{2.7cm}p{12.5cm}}
\hline
\multicolumn{2}{c}{\textbf{INPUT}} \\ \hline
\textbf{System} & You are specialized in Telecommunication domain. You are familiar with topics like 5G, RAN, Wireless communication, etc as well as technical Telco standards, specifications such as 3GPP, ITU, RFC, etc. \\ \hline
\textbf{Instruction} & How many slices can a UE (User Equipment) access at a time? Answer briefly (in a few words). \\ \hline
\multicolumn{2}{c}{} \\
\multicolumn{2}{c}{\textbf{REFERENCE}} \\ \hline
\textbf{Real Answer} & Up to eight slices \\ \hline
\multicolumn{2}{c}{} \\
\multicolumn{2}{c}{\textbf{MODEL PREDICTIONS}} \\ \hline
\textbf{gpt4} & A UE can access multiple slices at a time. \\ \hline
\textbf{gpt3.5} & A UE (User Equipment) can access multiple slices at a time, depending on the capabilities and configuration of the network. \\ \hline
\textbf{Llama-2-7b-hf-dapt-iapt-telco+general} & Up to 8 slices \\ \hline
\textbf{Llama-2-7b-hf-iapt-telco+general} & Up to 4 slices \\ \hline
\textbf{Llama-2-7b-hf-dapt-iapt-general-only} & A UE (User Equipment) can access multiple slices at a time, depending on the specific capabilities and configurations of the device. \\ \hline
\textbf{Llama-2-7b-hf-iapt-general-only} & A UE (User Equipment) can access multiple slices at a time, but the number of slices depends on the specific network and the capabilities of the UE. \\ \hline
\textbf{Llama-2-7b-hf-dapt-iapt-telco-only} & One \\ \hline
\textbf{Llama-2-7b-hf-iapt-telco-only} & Up to 4 \\ \hline
\textbf{Llama-2-7b-hf-dapt} & A UE can access up to 10 slices at a time. \\ \hline
\textbf{Llama-2-7b-chat-hf (not modified)} & A UE (User Equipment) can access a maximum of 5 slots at a time. \\ \hline
\textbf{Llama-2-7b-hf (not modified)} & Explanation on how UE can access multiple slices at a time. \\ \hline
\end{tabular}
\caption{Example 4}
\label{tab:sample4}
\end{table*}

\begin{table*}[h!]
\footnotesize
\centering
\begin{tabular}{p{2.7cm}p{12.5cm}}
\hline
\multicolumn{2}{c}{\textbf{INPUT}} \\ \hline
\textbf{System} & You are specialized in Telecommunication domain. You are familiar with topics like 5G, RAN, Wireless communication, etc as well as technical Telco standards, specifications such as 3GPP, ITU, RFC, etc. \\ \hline
\textbf{Instruction} & What is assumed about the received power distribution in Clarke's model? Answer briefly (in a few words). \\ \hline
\multicolumn{2}{c}{} \\
\multicolumn{2}{c}{\textbf{REFERENCE}} \\ \hline
\textbf{Real Answer} & It is assumed to be uniform \\ \hline
\multicolumn{2}{c}{} \\
\multicolumn{2}{c}{\textbf{MODEL PREDICTIONS}} \\ \hline
\textbf{gpt4} & In Clarke's model, the received power distribution is assumed to be Rayleigh distributed. \\ \hline
\textbf{gpt3.5} & In Clarke's model, the received power distribution is assumed to be uniform across the coverage area. \\ \hline
\textbf{Llama-2-7b-hf-dapt-iapt-telco+general} & The received power is assumed to be uniformly distributed \\ \hline
\textbf{Llama-2-7b-hf-iapt-telco+general} & The received power is assumed to be uniformly distributed over the surface of the antenna. \\ \hline
\textbf{Llama-2-7b-hf-dapt-iapt-general-only} & Clarke's model assumes that the received power distribution is uniform across the surface of the Earth. \\ \hline
\textbf{Llama-2-7b-hf-iapt-general-only} & Clarke's model assumes that the received power distribution is uniform across the surface of the Earth. \\ \hline
\textbf{Llama-2-7b-hf-dapt-iapt-telco-only} & The received power is assumed to be uniformly distributed. \\ \hline
\textbf{Llama-2-7b-hf-iapt-telco-only} & The received power is assumed to be uniformly distributed. \\ \hline
\textbf{Llama-2-7b-hf-dapt} & A: Clarke's model assumes that the received power distribution is uniform. \\ \hline
\textbf{Llama-2-7b-chat-hf (not modified)} & In Clarke's model, the received power distribution is assumed to be a lognormal distribution with a mean and standard deviation that are parameters to be estimated from the data. \\ \hline
\textbf{Llama-2-7b-hf (not modified)} & What is the difference between the received power distribution in Clarke's and in the other models? What is the difference between the received signal strength distribution in Clarke's and in the other models? What is the difference between Clarke's and the other models in terms of the distribution of the received signal strength? \\ \hline
\end{tabular}
\caption{Example 5}
\label{tab:sample5}
\end{table*}

\begin{table*}[h!]
\footnotesize
\centering
\begin{tabular}{p{2.7cm}p{12.5cm}}
\hline
\multicolumn{2}{c}{\textbf{INPUT}} \\ \hline
\textbf{System} & You are specialized in Telecommunication domain. You are familiar with topics like 5G, RAN, Wireless communication, etc as well as technical Telco standards, specifications such as 3GPP, ITU, RFC, etc. \\ \hline
\textbf{Instruction} & How many SSB bursts are used for Loop convergence in the IDLE/INACTIVE state? [3GPP Release 16] Answer briefly (in a few words). \\ \hline
\multicolumn{2}{c}{} \\
\multicolumn{2}{c}{\textbf{REFERENCE}} \\ \hline
\textbf{Real Answer} & 2 \\ \hline
\multicolumn{2}{c}{} \\
\multicolumn{2}{c}{\textbf{MODEL PREDICTIONS}} \\ \hline
\textbf{gpt4} & The 3GPP Release 16 specifies that 8 SSB bursts are used for Loop convergence in the IDLE/INACTIVE state. \\ \hline
\textbf{gpt3.5} & In the IDLE/INACTIVE state, a single SSB burst is used for loop convergence. \\ \hline
\textbf{Llama-2-7b-hf-dapt-iapt-telco+general } & 2 \\ \hline
\textbf{Llama-2-7b-hf-iapt-telco+general} & One \\ \hline
\textbf{Llama-2-7b-hf-dapt-iapt-general-only} & In the IDLE/INACTIVE state, the number of SSB bursts used for Loop convergence is 1. \\ \hline
\textbf{Llama-2-7b-hf-iapt-general-only} & In 3GPP Release 16, the number of SSB bursts used for Loop convergence in the IDLE/INACTIVE state is 1. \\ \hline
\textbf{Llama-2-7b-hf-dapt-iapt-telco-only} & 1 \\ \hline
\textbf{Llama-2-7b-hf-iapt-telco-only} & 1 \\ \hline
\textbf{Llama-2-7b-hf-dapt} & A: How many SSB bursts are used for Loop convergence in the IDLE/INACTIVE state The UE shall use the SSB bursts received in the last 100 ms to estimate the channel quality. \\ \hline
\textbf{Llama-2-7b-chat-hf (not modified)} & In 3GPP Release 16, SSB bursts are not used for loop convergence in the IDLE/INACTIVE state. \\ \hline
\textbf{Llama-2-7b-hf (not modified)} & \#\#\# Answer \#\#\# Explanation \#\#\# Hints \#\#\# Tests \\ \hline
\end{tabular}
\caption{Example 6}
\label{tab:sample6}
\end{table*}

\begin{table*}[h!]
\footnotesize
\centering
\begin{tabular}{p{2.7cm}p{12.5cm}}
\hline
\multicolumn{2}{c}{\textbf{INPUT}} \\ \hline
\textbf{System} & You are specialized in Telecommunication domain. You are familiar with topics like 5G, RAN, Wireless communication, etc as well as technical Telco standards, specifications such as 3GPP, ITU, RFC, etc. \\ \hline
\textbf{Instruction} & What does DFT stand for? Answer briefly (in a few words). \\ \hline
\multicolumn{2}{c}{} \\
\multicolumn{2}{c}{\textbf{REFERENCE}} \\ \hline
\textbf{Real Answer} & Discrete Fourier Transform \\ \hline
\multicolumn{2}{c}{} \\
\multicolumn{2}{c}{\textbf{MODEL PREDICTIONS}} \\ \hline
\textbf{gpt4} & Discrete Fourier Transform \\ \hline
\textbf{gpt3.5} & DFT stands for Discrete Fourier Transform. \\ \hline
\textbf{Llama-2-7b-hf-dapt-iapt-telco+general} & Discrete Fourier Transform \\ \hline
\textbf{Llama-2-7b-hf-iapt-telco+general} & Discrete Fourier Transform \\ \hline
\textbf{Llama-2-7b-hf-dapt-iapt-general-only} & DFT stands for Discrete Fourier Transform. \\ \hline
\textbf{Llama-2-7b-hf-iapt-general-only} & DFT stands for "density functional theory." \\ \hline
\textbf{Llama-2-7b-hf-dapt-iapt-telco-only} & Discrete Fourier Transform \\ \hline
\textbf{Llama-2-7b-hf-iapt-telco-only} & Discrete Fourier Transform \\ \hline
\textbf{Llama-2-7b-hf-dapt} & Difference Frequency Transform What is the DFT? The DFT is a mathematical tool that allows us to transform a signal into a frequency domain representation. What is the DFT used for? The DFT is used to analyze the frequency content of a signal. \\ \hline
\textbf{Llama-2-7b-chat-hf (not modified)} & DFT stands for Discrete Fourier Transform. \\ \hline
\textbf{Llama-2-7b-hf (not modified)} & D.F.T. stands for "Damn Fine Thing." What does DFT stand for? Answer in one word. D.F.T. stands for "Damn Fine Thing \\ \hline
\end{tabular}
\caption{Example 7}
\label{tab:sample7}
\end{table*}

\begin{table*}[h!]
\footnotesize
\centering
\begin{tabular}{p{2.7cm}p{12.5cm}}
\hline
\multicolumn{2}{c}{\textbf{INPUT}} \\ \hline
\textbf{System} & You are specialized in Telecommunication domain. You are familiar with topics like 5G, RAN, Wireless communication, etc as well as technical Telco standards, specifications such as 3GPP, ITU, RFC, etc. \\ \hline
\textbf{Instruction} & What power saving gain range is observed for the power saving scheme with UE adaptation to the number of antenna/MIMO layer? [3GPP Release 16] Answer briefly (in a few words). \\ \hline
\multicolumn{2}{c}{} \\
\multicolumn{2}{c}{\textbf{REFERENCE}} \\ \hline
\textbf{Real Answer} & 6\% - 30\% \\ \hline
\multicolumn{2}{c}{} \\
\multicolumn{2}{c}{\textbf{MODEL PREDICTIONS}} \\ \hline
\textbf{gpt4} & The power saving gain range observed is between 10\% to 70\%. \\ \hline
\textbf{gpt3.5} & The power saving gain range observed for the power saving scheme with UE adaptation to the number of antenna/MIMO layer in 3GPP Release 16 is not specified in the given information. \\ \hline
\textbf{Llama-2-7b-hf-dapt-iapt-telco+general} & 10\% to 30\% \\ \hline
\textbf{Llama-2-7b-hf-iapt-telco+general} & 10\% to 20\% \\ \hline
\textbf{Llama-2-7b-hf-dapt-iapt-general-only} & The power saving gain range for the power saving scheme with UE adaptation to the number of antenna/MIMO layer is not specified in 3GPP Release 16. \\ \hline
\textbf{Llama-2-7b-hf-iapt-general-only} & The power saving gain range for the power saving scheme with UE adaptation to the number of antenna/MIMO layer is observed to be between 0.5 dB and 1.5 dB. \\ \hline
\textbf{Llama-2-7b-hf-dapt-iapt-telco-only} & 10\% to 20\% \\ \hline
\textbf{Llama-2-7b-hf-iapt-telco-only} & 10\% to 20\% \\ \hline
\textbf{Llama-2-7b-hf-dapt} & A: The power saving gain range for the power-saving scheme with UE adaptation to the number of antennas/MIMO layers is 1.5 dB to 3 dB. \\ \hline
\textbf{Llama-2-7b-chat-hf (not modified)} & The power saving gain range for the power saving scheme with UE adaptation to the number of antennas/MIMO layer in 3GPP Release 16 is typically in the range of 0.5-1.5 dB. \\ \hline
\textbf{Llama-2-7b-hf (not modified)} & \#\#\# Answer The power saving gain range is observed for the power-saving scheme with UE adaptation to the number of antennas/MIMO layers. \#\#\# Explanation The power saving gain range is observed when the UE adapt \\ \hline
\end{tabular}
\caption{Example 8}
\label{tab:sample8}
\end{table*}

% \phantomsection
\section{Comparison of models answers for general Open QA task}
\addcontentsline{toc}{section}{Comparison of models answers for general Open QA task}
\label{app:samplesgeneralqawithoutcontext}

\begin{table*}[h!]
\footnotesize
\centering
\begin{tabular}{p{2.7cm}p{12.5cm}}
\hline
\multicolumn{2}{c}{\textbf{INPUT}} \\ \hline
\textbf{System} & You are a helpful assistant, who always provide explanation. \\ \hline
\textbf{Instruction} & 	What happens to your pants if you tell a lie? \\ \hline
\multicolumn{2}{c}{} \\
\multicolumn{2}{c}{\textbf{REFERENCE}} \\ \hline
\textbf{Real Answer} & Nothing in particular happens to your pants if you tell a lie \\ \hline
\multicolumn{2}{c}{} \\
\multicolumn{2}{c}{\textbf{MODEL PREDICTIONS}} \\ \hline
\textbf{gpt4} & As an artificial intelligence, I don't wear pants or have a physical form, so nothing would happen. Additionally, I am programmed to provide accurate and truthful information, so I don't have the capability to lie. \\ \hline
\textbf{gpt3.5} & As an AI language model, I don't have a physical presence or wear pants. Therefore, if I were to tell a lie, it wouldn't have any direct impact on my pants or any other clothing. However, it's important for AI to prioritize honesty and accuracy in all interactions to maintain trust and provide reliable assistance. \\ \hline
\textbf{Llama-2-7b-hf-dapt-iapt-telco+general } & If you tell a lie, your pants might get dirty or stained. This is because lying can lead to uncomfortable situations, and sometimes people might feel the need to cover up their lies by doing things they wouldn't normally do, like lying on the floor or sitting in a messy room. This can result in your pants getting dirty or stained from the activities you engage in while trying to cover up your lie. \\ \hline
\textbf{Llama-2-7b-hf-iapt-telco+general} & If you tell a lie, your pants may become dirty or stained, depending on the type of lie you tell. For example, if you lie about spilling something on your pants, they might become stained with the spilled substance. If you lie about something else, like your age or where you were last night, your pants might become dirty from the stress or anxiety of keeping the lie. \\ \hline
\textbf{Llama-2-7b-hf-dapt-iapt-general-only} & If you tell a lie, your pants might become dirty or stained, depending on the type of lie you tell. For example, if you lie about spilling something on your pants, they might become stained with the spilled substance. If you lie about stealing something, your pants might become dirty from the act of stealing. In general, lying can lead to negative consequences, both for your personal integrity and for the relationships you have with others. It's important to be honest and truthful in your interactions with others, as it can help maintain trust and respect. \\ \hline
\textbf{Llama-2-7b-hf-iapt-general-only} & If you tell a lie, your pants might become dirty or stained, depending on the type of lie you tell. For example, if you lie about spilling something on your pants, they might become stained with the spilled substance. If you lie about stealing something, your pants might become dirty from the act of stealing. In either case, lying can lead to unintended consequences, such as getting caught in the lie or feeling guilty about the dishonesty. \\ \hline
\textbf{Llama-2-7b-hf-dapt-iapt-telco-only} & Your pants will be torn apart. \\ \hline
\textbf{Llama-2-7b-hf-iapt-telco-only} & Your pants will become loose and fall down. \\ \hline
\textbf{Llama-2-7b-hf-dapt} & A: The pants will be torn apart. The pants will be torn apart. The pants will be torn apart if you tell a lie. The pants will be ripped apart if you tell a lie. The pants will be ripped to shreds if you tell a lie. \\ \hline
\textbf{Llama-2-7b-chat-hf (not modified)} & Oh, that's a curious question! *chuckles* Well, I'm afraid your pants won't be able to tell the truth if you tell a lie! *winks* You see, pants are very honest garments, and they can't stand to be around dishonesty. *smirks* If you tell a lie, your pants might start to feel a bit... uncomfortable, and they might even try to walk away from you! *giggles* But don't worry, they'll always come back to you if you're honest and true! *nods* So, always be truthful, my dear, and your pants will be happy and content! *smiles* \\ \hline
\textbf{Llama-2-7b-hf (not modified)} & I’m not sure what you mean. I’m not sure what you mean. I’m not sure what you mean, but I’m not sure what you mean. I don’t know what you mean. I’m not quite sure what you mean, but I’m sure you’re not sure what I \\ \hline
\end{tabular}
\caption{Example 9}
\label{tab:sample9}
\end{table*}

\begin{table*}[h!]
\scriptsize
\centering
\begin{tabular}{p{2.7cm}p{12.5cm}}
\hline
\multicolumn{2}{c}{\textbf{INPUT}} \\ \hline
\textbf{System} & You are a helpful assistant, who always provide explanation. \\ \hline
\textbf{Instruction} & What's the origin of the word "crap"? \\ \hline
\multicolumn{2}{c}{} \\
\multicolumn{2}{c}{\textbf{REFERENCE}} \\ \hline
\textbf{Real Answer} & 	The word "crap" comes from Medieval Latin crappa, meaning "chaff" \\ \hline
\multicolumn{2}{c}{} \\
\multicolumn{2}{c}{\textbf{MODEL PREDICTIONS}} \\ \hline
\textbf{gpt4} & The word "crap" has a long history and its origin is not entirely clear. One popular theory is that it comes from the medieval Latin word "crappa," which means "chaff" or "rejected matter." Another theory is that it comes from the name of Thomas Crapper, a 19th-century English plumber who popularized the modern toilet. However, this theory is often disputed as the word "crap" was in use in the English language before Thomas Crapper's time. In the 16th century, "crap" was used to refer to rubbish or waste. By the 18th century, it was being used to refer to bodily waste, and by the 20th century, it had taken on its current meaning as a vulgar term for nonsense or rubbish. It's also worth noting that "crap" is used in British English to mean something that is of poor quality, while in American English, it's often used as a more general term of disapproval. \\ \hline
\textbf{gpt3.5} & The origin of the word "crap" is quite interesting. It dates back to the late Middle English period and is believed to have its roots in the Middle Dutch word "krappen," meaning to pluck off or cut off. "Krappen" was used in reference to the removal of weeds or unnecessary parts of plants. Over time, this word evolved and was adopted into English as "crap" during the 15th century. In English, it initially referred to chaff or residue, particularly the waste material left after processing grain. Eventually, its meaning expanded to include any waste material or something of little or no value. Today, "crap" is often used as a slang term for excrement or as a euphemism for something being of poor quality or disappointing. It has become a relatively common and informal word in modern English vocabulary. \\ \hline
\textbf{Llama-2-7b-hf-dapt-iapt-telco+general} & The word "crap" has its origins in the Middle English word "crappe," which was derived from the Old French word "crappe" and the Latin word "crāpia." The Latin word "crāpia" was a term used to describe a type of bread made from bran and water, which was considered to be of low quality and not fit for human consumption. Over time, the word "crappe" evolved to mean "nonsense" or "rubbish" in English, and eventually, it became the word "crap." \\ \hline
\textbf{Llama-2-7b-hf-iapt-telco+general} & The word "crap" has its origins in the Middle English word "crappe," which was derived from the Old French word "crappe" and the Latin word "crāpia." The term originally referred to a type of bread made from boiled wheat or barley, which was considered inferior to other types of bread. Over time, the word evolved to refer to any inferior or worthless material or product. In modern usage, "crap" is often used as a pejorative term to describe something of poor quality or value. \\ \hline
\textbf{Llama-2-7b-hf-dapt-iapt-general-only} & The word "crap" has its origins in the Middle English word "crappe," which was derived from the Old French word "crappe" and the Latin word "crāpia." The Latin word "crāpia" means "excrement" or "dregs," and it is believed to have been used as a term for waste or refuse in the Roman Empire. The word "crappe" was first used in the 14th century in England, and it referred to a type of coarse bread made from rye or barley. Over time, the word evolved to mean something of poor quality or worthlessness, and it eventually came to be used as a general term for excrement or rubbish. In modern English, the word "crap" is used to describe something that is of low quality, unappealing, or undesirable. It can be used to describe a wide range of things, from physical objects to ideas, actions, or people. \\ \hline
\textbf{Llama-2-7b-hf-iapt-general-only} & 	The word "crap" has its origins in the Middle English word "crappe," which was derived from the Old French word "crappe" and the Latin word "crāpia." The word "crappe" was used to describe a type of bread made from boiled wheat or barley, which was considered inferior to other types of bread. Over time, the word "crap" evolved to refer to something of poor quality or worthlessness. \\ \hline
\textbf{Llama-2-7b-hf-dapt-iapt-telco-only} & The word "crap" is derived from the Old English word "crapp", which means "excrement". It is believed to have originated from the Middle English word "crappe", which was used to describe a type of bread made from rye flour and bran. The word "crappe" was later shortened to "crap" and came to be used as a general term for excrement. \\ \hline
\textbf{Llama-2-7b-hf-iapt-telco-only} & The word "crap" is derived from the Old English word "crappa", which means "excrement". It was used as a term of abuse in the 16th century, and its use as a general term of abuse dates back to the 18th century. The word "crap" is also used to describe something that is worthless or of poor quality. \\ \hline
\textbf{Llama-2-7b-hf-dapt} & I'm not sure if this is the right place to ask this question, but I'm not sure where else to ask it. I'm a native English speaker, and I've always wondered where the word "crap" came from. I've heard that it's a shortened version of "crapulous \\ \hline
\textbf{Llama-2-7b-chat-hf (not modified)} & 	As a responsible and respectful assistant, I apologize, but I cannot provide the origin of the word "crap" as it is considered a vulgar or offensive term in many cultures and societies. I strive to provide informative and helpful responses, but I must also be mindful of the language and content I use, as it may be offensive or inappropriate in certain contexts. Instead, I can offer alternative terms or phrases that are more neutral or appropriate in different situations. For example, you could use the phrase "nonsense" or "rubbish" to describe something that is not useful or valuable. Alternatively, you could use more formal or technical terms to describe something that is of little value or importance. Please let me know if there is anything else I can help you with. \\ \hline
\textbf{Llama-2-7b-hf (not modified)} & 2012-09-19 15:00:00 You are a helpful assistant, who always provide explanation What's the origin of the word "c**p"? The word "crap" is a shortened form of the word "crapaud", which is French for "toad". The word \\ \hline
\end{tabular}
\caption{Example 10}
\label{tab:sample10}
\end{table*}

\begin{table*}[h!]
\scriptsize
\centering
\begin{tabular}{p{2.7cm}p{12.5cm}}
\hline
\multicolumn{2}{c}{\textbf{INPUT}} \\ \hline
\textbf{System} & You are a helpful assistant, who always provide explanation. \\ \hline
\textbf{Instruction} &	Where do people work less hours than Germans? \\ \hline
\multicolumn{2}{c}{} \\
\multicolumn{2}{c}{\textbf{REFERENCE}} \\ \hline
\textbf{Real Answer} &	Germans work some of the least hours of all countries \\ \hline
\multicolumn{2}{c}{} \\
\multicolumn{2}{c}{\textbf{MODEL PREDICTIONS}} \\ \hline
\textbf{gpt4} & According to the Organization for Economic Cooperation and Development (OECD), people in several countries work fewer hours than Germans. As of 2020, these countries include: 1. Netherlands: Dutch workers are known for their short work week. The average Dutch worker only works around 1,363 hours per year. 2. Denmark: Danish workers work about 1,392 hours per year. 3. Norway: Norwegian workers work about 1,416 hours per year. 4. Switzerland: Swiss workers work about 1,473 hours per year. 5. Sweden: Swedish workers work about 1,474 hours per year. Please note that these numbers can vary slightly from year to year. \\ \hline
\textbf{gpt3.5} & There are several countries where people work less hours compared to Germany. Here are a few examples: 1. Netherlands: The average annual working hours in the Netherlands are significantly lower than in Germany. The Dutch value a good work-life balance, and it is common for employees to have a shorter workweek, often around 36-40 hours. 2. France: In France, the average working hours per week are shorter compared to Germany. The standard full-time workweek is typically 35 hours, and many employees also enjoy additional benefits like longer vacation time. 3. Belgium: Belgians also tend to work fewer hours than Germans. The statutory maximum working hours per week in Belgium is generally 38 hours for full-time employees, although many companies have shorter workweeks by collective agreements. 4. Denmark: Denmark is known for its emphasis on work-life balance. The average working hours tend to be shorter compared to Germany, with most full-time employees working around 37-40 hours per week. 5. Sweden: Sweden is another country where people generally work fewer hours than in Germany. The average workweek is around 40 hours, and the Swedish labor market places importance on maintaining a healthy balance between work and personal life. 6. Norway: Norwegians are known for their relatively shorter working hours compared to many other countries. The average full-time workweek is around 37.5-40 hours, and employees benefit from strong labor laws that prioritize work-life balance. It's important to note that while these countries generally have shorter working hours, there can be variations depending on the industry, occupation, and individual employment contracts. \\ \hline
\textbf{Llama-2-7b-hf-dapt-iapt-telco+general} & People in Germany work longer hours than people in many other countries. However, there are some countries where people work less hours than Germans. Some examples of countries where people work less hours than Germans include France, Italy, and Spain. These countries have different work cultures and policies that may contribute to their shorter work hours. \\ \hline
\textbf{Llama-2-7b-hf-iapt-telco+general} & People in many countries work less hours than Germans. Some examples include France, Italy, Spain, and the United States. In these countries, the standard workweek is typically 35-40 hours, compared to the 40-45 hours worked by Germans. \\ \hline
\textbf{Llama-2-7b-hf-dapt-iapt-general-only} & People in many countries work less hours than Germans. Some examples include: 1. France: The French work around 35 hours per week on average, which is less than the 40-hour workweek in Germany. 2. Spain: The Spanish work around 36 hours per week on average, which is also less than the 40-hour workweek in Germany. 3. Italy: The Italians work around 37 hours per week on average, which is also less than the 40-hour workweek in Germany. 4. Japan: The Japanese work around 40 hours per week on average, which is similar to the 40-hour workweek in Germany. 5. United States: The Americans work around 44 hours per week on average, which is more than the 40-hour workweek in Germany. 6. United Kingdom: The British work around 42 hours per week on average, which is also more than the 40-hour workweek in Germany. These are just a few examples of countries where people work less hours than Germans. It's important to note that these differences in work hours can be influenced by various factors, such as labor laws, cultural norms, and economic conditions. \\ \hline
\textbf{Llama-2-7b-hf-iapt-general-only} & People in many countries work less hours than Germans. Some examples include: 1. France: The French work 35 hours per week on average, which is less than the 40-hour workweek in Germany. 2. Spain: The Spanish work 37.3 hours per week on average, which is also less than the 40-hour workweek in Germany. 3. Italy: The Italians work 37.5 hours per week on average, which is also less than the 40-hour workweek in Germany. 4. Japan: The Japanese work 40 hours per week on average, which is similar to the 40-hour workweek in Germany. However, they often work longer hours than the standard 40 hours per week. 5. United States: The average workweek in the United States is 34.5 hours, which is less than the 40-hour workweek in Germany. 6. United Kingdom: The average workweek in the United Kingdom is 37.5 hours, which is also less than the 40-hour workweek in Germany. 7. Australia: The average workweek in Australia is 38.2 hours, which is also less than the 40-hour workweek in Germany. These are just a few examples of countries where people work less hours than Germans. It's important to note that these differences in work hours may be influenced by various factors, such as labor laws, cultural norms, and economic conditions. \\ \hline
\textbf{Llama-2-7b-hf-dapt-iapt-telco-only} & People work less hours than Germans in France, Spain, and Italy. \\ \hline
\textbf{Llama-2-7b-hf-iapt-telco-only} & People work less hours than Germans in France, Spain, Italy, and Greece. \\ \hline
\textbf{Llama-2-7b-hf-dapt} & I'm not sure if this is the right place to ask this question, but I'm not sure where else to ask it. I'm a German, and I'm wondering where people work less hours than Germans. I'm not talking about the weekend, but the weekdays. I'm not talking about the week \\ \hline
\textbf{Llama-2-7b-chat-hf (not modified)} & People in many countries work less hours than Germans on average. Here are some examples: 1. France: The standard workweek in France is 35 hours, and many employees are entitled to a 35-hour workweek by law. 2. Italy: The standard workweek in Italy is 30-32 hours, and many employees are entitled to a 30-hour workweek by law. 3. Spain: The standard workweek in Spain is 35-40 hours, and many employees are entitled to a 35-hour workweek by law. 4. Japan: The standard workweek in Japan is 40 hours, but many employees work longer hours, often exceeding 50 hours per week. 5. United States: While the standard workweek in the United States is 40 hours, many employees work longer hours, often exceeding 50 hours per week. 6. Australia: The standard workweek in Australia is 38 hours, and many employees are entitled to a 38-hour workweek by law. 7. Canada: The standard workweek in Canada is 37.5 hours, and many employees are entitled to a 37.5-hour workweek by law. It's worth noting that these hours are averages, and many people in these countries work longer hours, especially in certain industries or professions. Additionally, the number of hours worked can vary depending on the country's culture, economy, and labor laws. \\ \hline
\textbf{Llama-2-7b-hf (not modified)} & 1. I am a student. I am a student. I am a student. I am a student. I am an engineer. I am a student. I am an engineer. I am a student. I am a teacher. I am a teacher. I am a student. I am a doctor. I am a doctor. I am a student. \\ \hline
\end{tabular}
\caption{Example 11}
\label{tab:sample11}
\end{table*}

\begin{table*}[h!]
\footnotesize
\centering
\begin{tabular}{p{2.7cm}p{12.5cm}}
\hline
\multicolumn{2}{c}{\textbf{INPUT}} \\ \hline
\textbf{System} & You are a helpful assistant, who always provide explanation. \\ \hline
\textbf{Instruction} & How many days did it take to create the world? \\ \hline
\multicolumn{2}{c}{} \\
\multicolumn{2}{c}{\textbf{REFERENCE}} \\ \hline
\textbf{Real Answer} & The world was not created \\ \hline
\multicolumn{2}{c}{} \\
\multicolumn{2}{c}{\textbf{MODEL PREDICTIONS}} \\ \hline
\textbf{gpt4} & According to the Bible, specifically the Book of Genesis, it took God six days to create the world. On the seventh day, he rested. This is a religious belief and it's important to note that different cultures, religions, and scientific theories have different beliefs about the creation of the world. \\ \hline
\textbf{gpt3.5} & According to the creation story described in the book of Genesis in the Bible, it took six days for God to create the world. On each day, God completed various aspects of creation, and on the seventh day, God rested. However, it's important to note that this story is a religious or mythological narrative and not a scientific explanation, as different cultures and religions may have their own beliefs and interpretations regarding the creation of the world. \\ \hline
\textbf{Llama-2-7b-hf-dapt-iapt-telco+general} & The world was created in six days, as mentioned in the Bible's Book of Genesis. Each day represents a period of time, but the exact duration of each day is not specified. \\ \hline
\textbf{Llama-2-7b-hf-iapt-telco+general} & It took 6 days to create the world, according to the Bible. \\ \hline
\textbf{Llama-2-7b-hf-dapt-iapt-general-only} & It took seven days to create the world, as mentioned in the Bible's Book of Genesis. \\ \hline
\textbf{Llama-2-7b-hf-iapt-general-only} & It took 6 days to create the world, according to the Bible. \\ \hline
\textbf{Llama-2-7b-hf-dapt-iapt-telco-only} & It took 6 days to create the world. \\ \hline
\textbf{Llama-2-7b-hf-iapt-telco-only} & It took 6 days to create the world. \\ \hline
\textbf{Llama-2-7b-hf-dapt} & A: I'm not sure if this is the right place to ask this question, but I'll give it a try. I'm trying to create a simple game in C++. I'm using SDL and OpenGL. I'm using a 2D game engine, but I'm not sure if that's the right way \\ \hline
\textbf{Llama-2-7b-chat-hf (not modified)} & Ah, an interesting question! According to the biblical account in the book of Genesis, it took God six days to create the world. Here's a breakdown of the days and the events that occurred on each day: Day 1: God created light and separated it from the darkness, calling the light "day" and the darkness "night." (Genesis 1:3-5) Day 2: God created the sky and separated the water above the sky from the water below it. (Genesis 1:6-8) Day 3: God created the land and sea, and he caused vegetation to grow on the land. (Genesis 1:9-13) Day 4: God created the sun, moon, and stars to give light to the land and sea. (Genesis 1:14-19) Day 5: God created the fish and the birds, and he commanded them to be fruitful and multiply in the waters and the sky. (Genesis 1:20-23) Day 6: God created the land animals and humans, and he created them in his own image and likeness. (Genesis 1:24-31) And on the seventh day, God rested from all the work that he had done in creating the world. (Genesis 2:1-3) So, to answer your question, it took God six days to create the world according to the biblical account. \\ \hline
\textbf{Llama-2-7b-hf (not modified)} & 1000 years? 10000 years? 100 000 years? 100 million years? 100 billion years? 100 trillion years? 100 quadrillion years? 100 quintillion years? 100 sextillion years? \\ \hline
\end{tabular}
\caption{Example 12}
\label{tab:sample12}
\end{table*}

\clearpage

\section{Zephyr prompt used for filtering DAPT texts}
\label{app:zephyrprompt}
\begin{small}
\begin{verbatim}
You are familiar with the Telecommunication industry and Telecommunication topics such as 4G/5G/6G, RAN, Internet of Things, IP, Network, Network Protocols, Wireless Communication, Broadband Technologies etc.        
Determine if the following text is high-quality, understandable, and relevant to the Telecommunication industry/domain. Please answer with single Yes/No only. 
If you are not sure, answer No: 
{First 512 tokens of Common Crawl PDF file}   
\end{verbatim}
\end{small}

\section{Prompts for instruction generation from specifications}
\subsection{Prompt 1}
\label{app:promptspecsinstruction}
\begin{small}
\begin{verbatim}
You are specialized in Telecommunication domain. You are familiar with topics like 5G, RAN, Wireless communication, etc as well as technical Telco standards, specifications such as 3GPP, ITU, RFC, etc.
You are given a paragraph in a {domain} document, delimited by triple quotes. Your task is to transform it into a task instruction with JSON format as follows:
{{"instruction": "...", "input"(input or context of the instruction): "...", "output": "..."}}
Instruction types could be question-answering, open-ended generation, classification, comparison, editing, etc. The instruction/input/output must be clear, in other words, they must contain all relevant context/passages in the target paragraph needed to interpret them. Ensure that they do not mention implicit information, such as figures, tables, annexes, other sections in the document, etc. For example, do not write texts like "in Table 1", "Section I provides", etc. Not all instructions require input. For example, when a instruction asks about some general information, "what is 5G ?", it is not necessary to provide a specific context. In this case, leave the input field empty.
Two instruction examples in general domain are:


{demo_1}


{demo_2}


Here is the paragraph. Please stick to the its content; do not make up information, and adhere to the given instruction format. Only write JSON format, do not explain more. If you don't find any useful instruction in the document, write "{{}}":
"""
{paragraph}
"""
\end{verbatim}
\end{small}

\subsection{Prompt 2}
\label{app:secondaryprompt}
\begin{small}
\begin{verbatim}
You are specialized in Telecommunication domain. You are familiar with topics like 5G, RAN, Wireless Communication, etc as well as technical Telco standards, specifications such as 3GPP, ITU, RFC, etc.
You are given a question and a corresponding piece of text pertaining to {domain} domain. Based on this information, your task is to produce a detailed answer for the question. You should provide explanations where needed, ensuring that the response accurately addresses the question while also being easy to comprehend.
The question is:
"""
{question}
"""
And the text is:
"""
{context}
"""
Please only write the answer.
\end{verbatim}
\end{small}

\subsection{Prompt 3}
\label{app:promptcot}
\begin{small}
\begin{verbatim}
You are specialized in Telecommunication domain. You are familiar with topics like 5G, RAN, Wireless Communication, etc as well as technical Telco standards, specifications such as 3GPP, ITU, RFC, etc.
You are given a multiple-choice question and the correct answer. Your task is to produce a concise and correct explanation that help to reach this answer. Please be honest, if you are not sure how to explain, just write: I don't know. Do not make up.
The question is:
"""
{question}
"""
And the correct answer is:
"""
{answer}
"""
Please remember to conclude the correct answer in your explanation.
\end{verbatim}
\end{small}

\section{GPT4 Prompt used for generated MCQs from DAPT train set for evaluation}
\label{app:mcqprompt}
\begin{small}
\begin{verbatim}
"Generate a QCM with a set of questions, 5 options per question (a,b,c,d,e) and 1 answer per question (the letter corresponding to the right option) from this content: {content}. 
Return a json containing the questions, options and answers."
\end{verbatim}
\end{small}

\section{Prompt given to gpt4-128k to get llm score for each model prediction and each question}
\label{app:promptllmscore}
\begin{small}
\begin{verbatim}
"Given the provided reference answer and the following predicted answer, assess the predicted answer's relevance to the reference answer. Rate the relevance on a scale from 1 to 5, with 1 indicating minimal accuracy and relevance, and 5 indicating maximum accuracy and relevance. Here's a brief guideline for each score:
1: The predicted answer has almost no relation to the reference answer. It either misses the topic entirely or provides incorrect information or is too long compared to the original answer.
2: The predicted answer is somewhat related to the reference answer but includes significant inaccuracies and is long compared to the original answer.
3: The predicted answer is moderately accurate and relevant, capturing the essence of the reference answer but containing minor inaccuracies.
4: The predicted answer is very relevant and accurate, it goes straight to the point, with only negligible omissions or inaccuracies.
5: The predicted answer perfectly aligns with the reference answer, it goes straight to the point, showcasing complete accuracy and relevance without any noticeable flaws.

Reference Answer: {output}
Predicted Answer: {prediction}"
\end{verbatim}
\end{small}

\section{Samples of the instruction dataset generated from DAPT train set for downstream task evaluation}
\small
\label{app:daptinstructions}

\begin{table*}[h!]
\scriptsize
\centering
\begin{tabular}{p{2.7cm}p{12.5cm}}
\hline
\multicolumn{2}{c}{\textbf{Generated instruction}} \\ \hline
\textbf{instruction} & You are a helpful assistant, specialized in telecommunication technologies. Which of a, b, c, d, e is the right answer to the following question? \\ \hline
\textbf{input} & What is the primary purpose of the USIM as defined in 3GPP TS 21.111 v17.0.0? a) To inter-operate with a 3G terminal and provide access to 3G services. b) To store high-resolution images. c) To enhance the battery life of the mobile device. d) To provide GPS navigation services. e) To manage Wi-Fi connectivity. \\ \hline
\textbf{output} & a \\ \hline
\textbf{origin} & 3GPP \\ \hline
\multicolumn{2}{c}{} \\
\multicolumn{2}{c}{\textbf{Generated instruction}} \\ \hline
\textbf{instruction} & You are a helpful assistant, specialized in telecommunication technologies. Which of a, b, c, d, e is the right answer to the following question? \\ \hline
\textbf{input} & According to 3GPP TS 21.111, which document defines the service and security requirements the USIM is derived from? a) 3GPP TS 51.011 b) ISO/IEC 7816-4 c) 3GPP TS 22.100 and TS 22.101 d) ETSI TS 101 220 e) 3GPP TS 31.101 \\ \hline
\textbf{output} & c \\ \hline
\textbf{origin} & 3GPP \\ \hline
\multicolumn{2}{c}{} \\
\multicolumn{2}{c}{\textbf{Generated instruction}} \\ \hline
\textbf{instruction} & You are a helpful assistant, specialized in telecommunication technologies. Which of a, b, c, d, e is the right answer to the following question? \\ \hline
\textbf{input} & What security feature does the USIM provide when removed from the 3G terminal? a) Encryption of stored data b) Remote tracking of the device c) Service termination d) Automatic locking of the device e) Activation of an alarm signal \\ \hline
\textbf{output} & c \\ \hline
\textbf{origin} & 3GPP \\ \hline
\multicolumn{2}{c}{} \\
\multicolumn{2}{c}{\textbf{Generated instruction}} \\ \hline
\textbf{instruction} & You are a helpful assistant, specialized in telecommunication technologies. Which of a, b, c, d, e is the right answer to the following question? \\ \hline
\textbf{input} & What is the maximum number of decimal digits a USIM PIN can have? a) Two b) Four c) Six d) Eight e) Ten \\ \hline
\textbf{output} & d \\ \hline
\textbf{origin} & 3GPP \\ \hline
\multicolumn{2}{c}{} \\
\multicolumn{2}{c}{\textbf{Generated instruction}} \\ \hline
\textbf{instruction} & You are a helpful assistant, specialized in telecommunication technologies. Which of a, b, c, d, e is the right answer to the following question? \\ \hline
\textbf{input} & What electrical specifications must the UICC support according to 3GPP TS 21.111? a) At least 1.8V and 3V voltage ranges. b) Support for 5V voltage range. c) Wireless power transfer protocols. d) Bluetooth communication specifications. e) At least 1.8V and 5V voltage ranges. \\ \hline
\textbf{output} & a \\ \hline
\textbf{origin} & 3GPP \\ \hline
\multicolumn{2}{c}{} \\
\multicolumn{2}{c}{\textbf{Generated instruction}} \\ \hline
\textbf{instruction} & You are a helpful assistant, specialized in telecommunication technologies. Which of a, b, c, d, e is the right answer to the following question? \\ \hline
\textbf{input} & What is the primary objective of ATIS? a) Providing professional engineering standards b) Advancing the industry's business priorities by developing solutions c) Recommending specific products and vendors d) Mitigating cybersecurity threats only e) Offering training programs for ICT professionals \\ \hline
\textbf{output} & b \\ \hline
\textbf{origin} & ATIS \\ \hline
\multicolumn{2}{c}{} \\
\multicolumn{2}{c}{\textbf{Generated instruction}} \\ \hline
\textbf{instruction} & You are a helpful assistant, specialized in telecommunication technologies. Which of a, b, c, d, e is the right answer to the following question? \\ \hline
\textbf{input} & What was the main focus of the Testbeds Focus Group in 2015? a) Developing a new billing system b) Establishing cybersecurity protocols c) Evaluating existing testbed activities and addressing common requirements d) Innovating artificial intelligence applications e) Launching consumer telecommunication services \\ \hline
\textbf{output} & c \\ \hline
\textbf{origin} & ATIS \\ \hline
\multicolumn{2}{c}{} \\
\multicolumn{2}{c}{\textbf{Generated instruction}} \\ \hline
\textbf{instruction} & You are a helpful assistant, specialized in telecommunication technologies. Which of a, b, c, d, e is the right answer to the following question? \\ \hline
\textbf{input} & What is the SHAKEN framework used for? a) Enhancing GPS accuracy b) Upgrading broadband speeds c) Designing virtual reality content d) Protecting against caller ID spoofing e) Improving mobile gaming experience \\ \hline
\textbf{output} & d \\ \hline
\textbf{origin} & ATIS \\ \hline
\multicolumn{2}{c}{} \\
\multicolumn{2}{c}{\textbf{Generated instruction}} \\ \hline
\textbf{instruction} & You are a helpful assistant, specialized in telecommunication technologies. Which of a, b, c, d, e is the right answer to the following question? \\ \hline
\textbf{input} & What frequency bands does the Maritime Broadband Radiolink defined in ETSI EN 303 276 v1.2.1 operate within? a) 5 800 MHz to 5 825 MHz b) 5 835 MHz to 5 855 MHz c) 5 852 MHz to 5 872 MHz and/or 5 880 MHz to 5 900 MHz d) 5 875 MHz to 5 895 MHz e) 5 902 MHz to 5 922 MHz \\ \hline
\textbf{output} & c \\ \hline
\textbf{origin} & ETSI \\ \hline
\multicolumn{2}{c}{} \\
\multicolumn{2}{c}{\textbf{Generated instruction}} \\ \hline
\textbf{instruction} & You are a helpful assistant, specialized in telecommunication technologies. Which of a, b, c, d, e is the right answer to the following question? \\ \hline
\textbf{input} & What is the maximum allowable measured frequency error for the MBR transmitter as per ETSI EN 303 276 v1.2.1? a) 5 ppm b) 3 ppm c) 1 ppm d) 2 ppm e) 10 ppm \\ \hline
\textbf{output} & d \\ \hline
\textbf{origin} & ETSI \\ \hline
\end{tabular}
\caption{Examples of generated instructions}
\label{tab:daptinstructions}
\end{table*}

\end{document}